\documentclass[10pt,twocolumn,letterpaper]{article}
\pdfoutput=1 

\usepackage{cvpr}
\usepackage{times}
\usepackage{epsfig}
\usepackage{graphicx}
\usepackage{amsmath}
\usepackage{amssymb}

\usepackage{multirow}

\usepackage[pagebackref=true,breaklinks=true,letterpaper=true,colorlinks,bookmarks=false]{hyperref}

\cvprfinalcopy 

\def\cvprPaperID{131} 

\newcommand{\Width}{\mathit{W}}
\newcommand{\Height}{\mathit{H}}

\newcommand{\Pose}{\mathbf{p}}
\newcommand{\JointProb}{\mathbf{v}}
\newcommand{\Image}{\mathbf{I}}
\newcommand{\NJoints}{\mathnormal{N_J}}
\newcommand{\NFeat}{\mathnormal{N_f}}
\newcommand{\Dim}{\mathnormal{D}}
\newcommand{\K}{\mathnormal{K}}
\newcommand{\T}{\mathnormal{T}}
\newcommand{\F}{\mathnormal{F}}
\newcommand{\M}{\mathnormal{M}}
\newcommand{\V}{\mathnormal{V}}
\newcommand{\Loss}{\mathit{L}}

\newcommand{\x}{\mathbf{x}}

\begin{document}

\title{
  2D/3D Pose Estimation and Action Recognition using Multitask Deep Learning
}

\author{Diogo C. Luvizon$^1$, David Picard$^{1,2}$, Hedi Tabia$^1$\\
$^1$ETIS UMR 8051, Paris Seine University, ENSEA, CNRS, F-95000, Cergy, France \\
$^2$Sorbonne Universit{\'e}, CNRS, Laboratoire d'Informatique de Paris 6, LIP6, F-75005
Paris, France \\
{\tt\small \{diogo.luvizon,  picard, hedi.tabia\}@ensea.fr}\\
}

\maketitle

\begin{abstract}
  %
  %
  Action recognition and human pose estimation are closely related
  but both problems are generally handled as distinct tasks in the literature.
  In this work, we propose a multitask framework for jointly 2D and 3D pose
  estimation from still images and human action recognition from video
  sequences.
  We show that a single architecture can be used to solve the two problems in an
  efficient way and still achieves state-of-the-art results.
  Additionally, we demonstrate that optimization from end-to-end leads to
  significantly higher accuracy than separated learning.
  The proposed architecture can be trained with data from different categories
  simultaneously in a seamlessly way.
  The reported results on four datasets (MPII, Human3.6M, Penn Action and NTU)
  demonstrate the effectiveness of our method on the targeted tasks.
  %
\end{abstract}




\section{Introduction}

\begin{figure}[htbp]
  \centering
  \includegraphics[width=7.75cm]{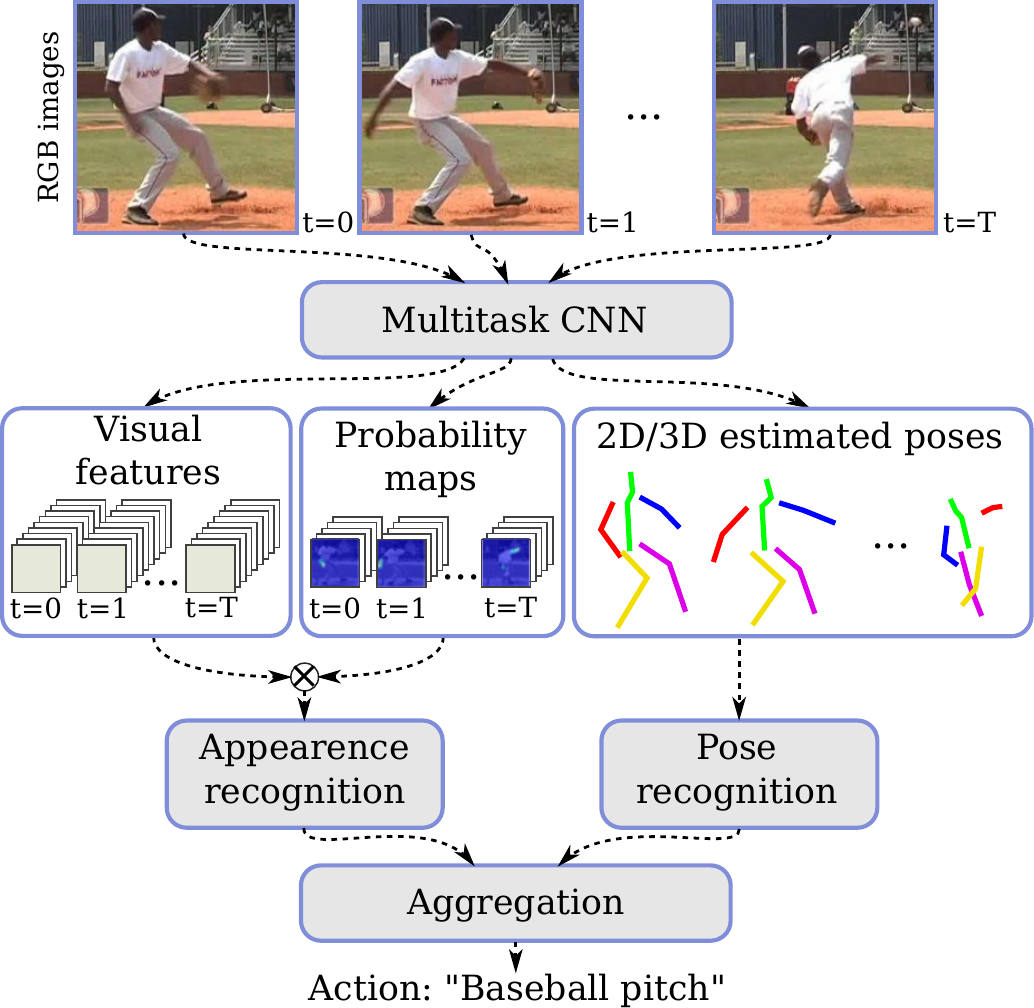}
  \caption{
    The proposed multitask approach for pose estimation and action
    recognition.  Our method provides 2D/3D pose estimation from single
    images or frame sequences. Pose and visual information are used
    to predict actions in a unified framework.
  }
  \label{fig:intro}
\end{figure}

Human action recognition and pose estimation have received an important
attention in the last years, not only because of their many applications, such
as video surveillance and human-computer interfaces, but also because they are
still challenging tasks.
%
%
Pose estimation and action recognition are usually handled as distinct
problems~\cite{cheronICCV15} or the last is used as a prior for the
first~\cite{Yao2012, Iqbal_2017}.
Despite the fact that pose is of extreme relevance for action recognition,
to the best of our knowledge, there is no method in the literature that
solves both problems in a joint way to the benefit of action recognition.
In that direction, our work proposes unique end-to-end trainable multitask
framework to handle 2D and 3D human pose estimation and action recognition
jointly, as presented in \autoref{fig:intro}.

One of the major advantages of deep learning is its capability to perform
end-to-end optimization.  As suggested by Kokkinos~\cite{Kokkinos17cvpr}, this
is all the more true for multitask problems, where related tasks can benefit
from one another.
Recent methods based on deep convolutional neural networks (CNNs) have achieved
impressive results on both 2D and 3D pose estimation tasks thanks to the rise
of new architectures and the availability of large amounts of
data~\cite{Newell_ECCV_2016, Pavlakos_2017_CVPR}.
Similarly, action recognition has recently been improved by using deep neural
networks relying on human pose \cite{baradel2017a}.  We believe both tasks have
not yet been stitched together to perform a beneficial joint optimization
because most pose estimation methods perform heat map prediction.  These
detection based approaches require the non-differentiable \textit{argmax}
function to recover the joint coordinates as a post processing stage, which
breaks the backpropagation chain needed for end-to-end learning.
%
%

We propose to solve this problem by extending the differentiable
Soft-argmax~\cite{Luvizon_2017_CoRR, Yi_2016} for joint 2D and 3D pose
estimation.  This allows us to stack action recognition on top of pose
estimation, resulting in a multitask framework trainable from end-to-end.
We present our contributions as follows:
\textbf{First}, the proposed pose estimation method achieves state-of-the-art
results on 3D pose estimation and the most accurate results among regression
methods for 2D pose estimation.
\textbf{Second}, the proposed pose estimation method is based on still images,
so it benefits from images ``in the wild'' for both 2D and 3D predictions.
This have been proven a very efficient way to learn visual features, which is also
very important for action recognition.
\textbf{Third}, our action recognition approach is based only on RGB images, from which
we extract pose and visual information. Despite that, we reached state-of-the-art
results on both 2D and 3D scenarios, even when compared with methods using ground-truth poses.
\textbf{Fourth}, the pose estimation method can be trained with multiple types of datasets
simultaneously, which makes it able to generalize 3D predictions from 2D annotated data.

The rest of this paper is organized as follows. In section~\ref{sec:relatedwork}
we present a review of the related work. The proposed framework is presented in
sections~\ref{sec:pose-estimation} and \ref{sec:action-recognition}, respectively
for the regression method for pose estimation and human action recognition.
Our extensive experiments are shown in section~\ref{sec:experiments}, followed by
our conclusions in section~\ref{sec:conclusions}.

\section{Related work}
\label{sec:relatedwork}

In this section, we present some of the most relevant methods to our work,
which are divided into \textit{human pose estimation} and \textit{action
recognition}.
Since an extensive literature review is prohibitive here due to the limited
size of the paper, we encourage the readers to refer to the surveys
in~\cite{SARAFIANOS20161, Herath_2017} for respectively pose estimation and
action recognition.

\subsection{Human pose estimation}

\textbf{2D pose estimation.}
The problem of human pose estimation has been intensively studied in the last
years, from Pictorial Structures~\cite{Andriluka_CVPR_2009,
Dantone_CVPR_2013, Pishchulin_CVPR_2013} to more recent CNN
approaches~\cite{Ning2017, Lifshitz_ECCV_2016, Pishchulin_CVPR_2016,%
Insafutdinov_ECCV_2016, Rafi_BMVC_2016, Wei_CVPR_2016,%
Belagiannis_ICCV_2015, Tompson_CVPR_2015, Toshev_CVPR_2014, pfister2014deep}.
From the literature, we can see that there are two distinct families of methods
for pose estimation: detection based and regression based methods.
Detection based methods handle pose estimation as a heat map prediction
problem, where each pixel in a heat map represents the detection score
of a corresponding joint~\cite{Bulat_ECCV_2016, Gkioxari_ECCV_2016}.
Exploring the concepts of stacked architectures, residual 
connections, and multiscale processing, Newell
\etal~\cite{Newell_ECCV_2016} proposed
the Stacked Hourglass Network, which improved scores on 2D pose
estimation challenges significantly.
Since then, methods in the state of the art are proposing complex variations
of the Stacked Hourglass architecture.
For example, Chu \etal~\cite{Chu_CVPR_17} proposed an attention model based
on conditional random field (CRF) and Yang \etal~\cite{Yang_2017_ICCV}
replaced the residual unit by a Pyramid Residual Module (PRM).
Generative Adversarial Networks (GANs) have been used to improve the capacity
of learning structural information~\cite{Chen_2017_ICCV} as well as to
refine the heat maps by learning more plausible predictions~\cite{ChouCC17},

However, detection approaches do not provide joint coordinates directly.
To recover the pose in $(x, y)$ coordinates, the \textit{argmax} function
is usually applied as a post-processing step.
On the other hand, regression based approaches use a nonlinear function that
maps the input directly to the desired output, which can be the joint coordinates.
Following this paradigm, Toshev and Szegedy~\cite{Toshev_CVPR_2014} proposed
a holistic solution based on cascade regression for body part detection and
Carreira \etal~\cite{Carreira_CVPR_2016} proposed the Iterative Error
Feedback.
The limitation of regression methods is that the regression function is
frequently sub-optimal. In order to tackle this weakness, the
Soft-argmax function~\cite{Luvizon_2017_CoRR} has been proposed to convert
heat maps directly to joint coordinates and consequently allow detection methods to be transformed into regression methods.
The main advantage of regression methods over detection ones is that they often are fully differentiable.
This means that the output of the pose estimation can be used in further processing and the whole system can be fine-tuned.

\textbf{3D pose estimation.}
%
Recently, deep architectures have been used to learn precise 3D representations
from RGB
images~\cite{Zhou_2017, Tome_2017_CVPR, Martinez_2017, Tekin_2016,
MehtaRCSXT16, Popa_2017_CVPR}, thanks to the availability of high quality
data~\cite{h36m_pami}, and are now able to surpass
depth-sensors~\cite{VNect_SIGGRAPH2017}.
Chen and Ramanan~\cite{Chen_2017_CVPR} divided the problem of 3D pose estimation
into two parts. First, they handle the 2D pose estimation considering the
camera coordinates and second, the estimated poses are matched to 3D
representations by means of a nonparametric shape model.
A bone representation of the human pose was proposed to reduce the data
variance~\cite{Sun_2017_ICCV}, however, such a structural transformation might effect
negatively tasks that depend on the extremities of the human body, since the error
is accumulated as we go away from the root joint.
Pavlakos \etal~\cite{Pavlakos_2017_CVPR} proposed the volumetric stacked
hourglass architecture.  However, the method suffers from the significant increase
in the number of parameters and in the required memory to store all the gradients.
In our approach, we also propose an intermediate volumetric representation for 3D
poses, but we use a much lower resolution than in~\cite{Pavlakos_2017_CVPR} and still
are able to increase significantly the state-of-the-art results, since our method
is based on a continuous regression function.

\subsection{Action recognition}

\textbf{2D action recognition.}
Action recognition from videos is considered a difficult problem because it
involves high level abstraction, and furthermore the temporal dimension
is not easily handled.
Previous approaches have explored classical methods for
features extraction~\cite{Nie_2015_CVPR, Jhuang_2013_ICCV}, where the
key idea is to use body joint locations to select visual features in space and time.
3D convolutions have been stated recently as the option that gives the highest
classification scores~\cite{Cao_2017, Carreira_2017_CVPR, varol17a},
but they involve high number of parameters, require an elevated
amount of memory for training, and cannot efficiently benefit from the abundant
still images for training.
Action recognition is improved by attention models that focus on body
parts~\cite{Baradel_CVPR_2018} and two-stream networks can be used to merge
both RGB images and the costly optical flow maps~\cite{cheronICCV15}.
%
%
%

Most 2D action recognition methods use the body joint information
only to extract localized visual features, as an attention mechanism.
The few methods that directly explore the body joints do not generate it, therefore
they are limited to datasets that provide skeletal data.
Our approach removes these limitations by performing pose estimation together with action recognition.
As such, our model only needs the input RGB frames while still performing discriminative visual recognition guided by estimated body joints.

\textbf{3D action recognition.}
Differently from video based action recognition, 3D action recognition is mostly based on skeleton data
as the primary information~\cite{Luvizon_PRL_2017, Presti20163DSH}.
With recently available depth sensors such as the Microsoft Kinect, it is possible
to capture 3D skeletal data without a complex installation
procedure frequently required for motion capture systems (MoCap).
However, due to the use of infrared projectors, these depth sensors are limited to indoor environments.
Moreover, they have a low range precision and are not robust to occlusions, frequently
resulting in noisy skeletons.

To cope with noisy skeletons, Spatio-Temporal LSTM networks have been widely used
by applying a gating mechanism~\cite{Liu2016} to learn the reliability of
skeleton sequences or by using attention
mechanisms~\cite{Liu_2017_CVPR, Song_2017_AAAI}.
In addition to the skeleton data, multimodal approaches can also benefit
from the visual cues~\cite{Shahroudy2017DeepMF}.
In that direction, Baradel \etal~\cite{baradel2017a} proposed the
Pose-conditioned Spatio-Temporal attention mechanism by using the skeleton
sequences for both spatial and temporal attention mechanisms,
while action classification is based on pose and appearance features
extracted from patches on the hands.


Since our architecture predicts high precision 3D skeleton from the input RGB frames,
we do not have to cope with the noisy skeletons from Kinect.
Moreover, we show in the experiments that, despite being based on temporal convolution instead of the more common LSTM, our system is able to reach state of the art performance on 3D action recognition.

\section{Human pose estimation}
\label{sec:pose-estimation}

Our approach for human pose estimation is a regression method, similarly
to~\cite{Luvizon_2017_CoRR, Sun_2017_ICCV, Carreira_CVPR_2016}.
We extended the Soft-argmax function to handle 2D and 3D pose regression
in a unified way.
The details of our approach are explained as follows.

\subsection{Regression-based approach}


The human pose regression problem is defined by the input RGB image
$\Image{}\in\mathbb{R}^{\Width{}\times\Height{}\times3}$, the output
estimated pose $\hat{\Pose} \in\mathbb{R}^{\NJoints{}\times\Dim{}}$
with $\NJoints{}$ body joints of dimension $\Dim{}$, and
a regression function $f_{r}$, as given by the following equation:
\begin{equation}
    \hat{\Pose{}} = f_{r}(\Image{}, \theta_{r}),
\end{equation}
where $\theta_{r}$ is a set of trainable parameters of function $f_r$.
The objective is to optimize the parameters $\theta_{r}$ in order to minimize
the error between the estimated pose $\hat{\Pose{}}$ and the ground truth pose
$\Pose{}$.
In order to implement this function, we use a deep CNN.
As the pose estimation is the first part of our multitask approach, the function
$f_{r}$ has to be differentiable in order to allow end-to-end optimization.
This is made possible by the Soft-argmax,
which is a differentiable alternative to the \textit{argmax} function
and can be used to convert heat maps to $(x,y)$ joint coordinates proposed in \cite{Luvizon_2017_CoRR}.

\subsubsection{Network architecture}
\begin{figure}[htbp]
  \centering
  \includegraphics[width=8.00cm]{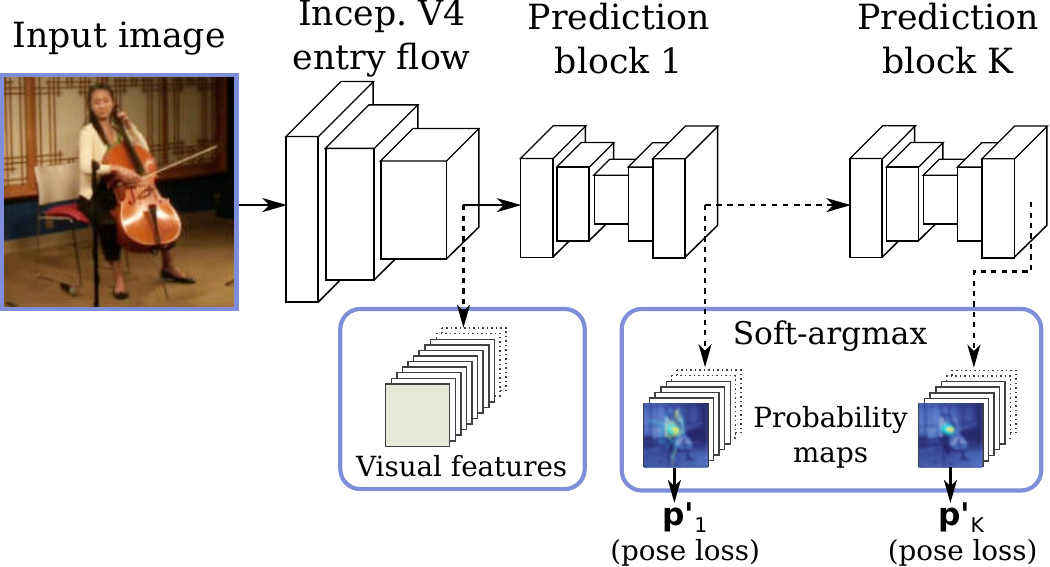}
  \caption{
    Human pose regression approach from a single RGB frame.
    The input image is fed through a CNN composed by one
    entry flow and $\K{}$ prediction blocks. Predictions are refined at
    each prediction block.
  }
  \label{fig:poseregression-arch}
\end{figure}

The network architecture has its entry flow based on
Inception-V4~\cite{SzegedyIV16} that is used to provide basic features
extraction. Then, similarly to what is found in \cite{Luvizon_2017_CoRR}, $\K{}$ prediction
blocks are used to refine estimations, from which we use the last prediction
$\Pose{}'_K$ as our estimated pose $\hat{\Pose{}}$.
Each prediction block is composed of eight residual depth-wise convolutions
separated into three different resolutions.
As a byproduct, we also have access to low-level visual features and to the
intermediate joint probability maps
that are indirectly learned thanks to the Soft-argmax layer.
In our method for action recognition, both visual features and joint
probability maps are used to produce appearance features, as detailed in
section~\ref{sec:visual}.
A graphical representation of the pose regression network is shown in
\autoref{fig:poseregression-arch}.

\subsubsection{The Soft-argmax layer}
\begin{figure}[htbp]
  \centering
  \includegraphics[width=7.75cm]{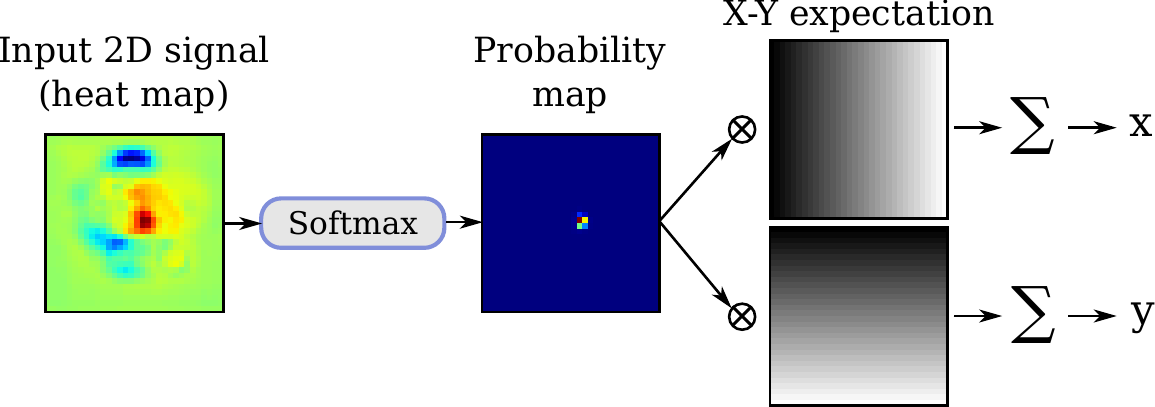}
  \caption{
    Graphical representation of the Soft-argmax operation for
    2D input signals (heat maps). The outputs are the coordinates $x$ and $y$
    that approximates the maximum in the input signal.
  }
  \label{fig:softmax}
\end{figure}

An intuitive graphical explanation of the Soft-argmax layer is shown
in \autoref{fig:softmax}.
Given an input signal, the main idea is to consider that the argument of the
maximum $\mu$ can be approximated by the expectation of the input signal after
being normalized to have the properties of a distribution.  Indeed, for a
sufficiently pointy (leptokurtic) distribution, the expectation should be close
to the maximum a posteriori (MAP) estimation.
The normalized exponential function (Softmax) is used, since it alleviates the
undesirable influences of values bellow the maximum and increases the
``pointiness'' of the resulting distribution.
For a 2D heat map as input, the normalized signal can be interpreted as
the \textit{probability map} of a joint being at position $(x, y)$, and the
expected value for the joint position is given by the expectation on
the normalized signal:

\begin{equation}
    \Psi(\x{}) = \left(\sum_{c=0}^{W_{\x{}}}\sum_{l=0}^{H_{\x{}}}\frac{c}{W_{\x{}}}\Phi(\x{})_{l,c},\\
    \sum_{c=0}^{W_{\x{}}}\sum_{l=0}^{H_{\x{}}}\frac{l}{H_{\x{}}}\Phi(\x{})_{l,c}\right)^{T},
\end{equation}
where $\x{}$ is the input heat map with dimension $W_{\x{}}\times{H_{\x{}}}$
and $\Phi$ is the Softmax normalization function.

\subsubsection{Joint visibility}
The probability of a certain joint being visible in the image
is computed by the Sigmoid function on the maximum value in the corresponding
input heat map.
Considering a pose layout with $\NJoints{}$ joints, the joint visibility
vector is represented by $\JointProb{}\in\mathbb{R}^{\NJoints{}\times{1}}$.
Remark that the \textit{visibility} information is unrelated to the joint
\textit{probability map}, since the latter always sums to one.


%
%

\subsection{Unified 2D/3D pose estimation}

\begin{figure}[htbp]
  \centering
  \includegraphics[width=7.75cm]{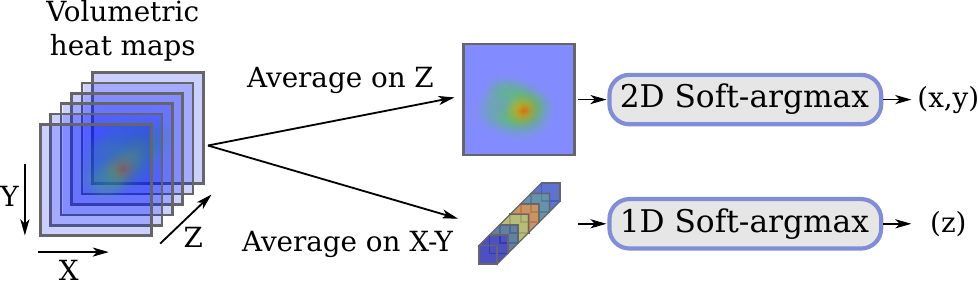}
  \caption{
    Unified 2D/3D pose estimation by using volumetric heat maps.
  }
  \label{fig:3dpose}
\end{figure}

We extended the 2D pose regression to 3D scenarios by expanding 2D heat maps
to volumetric representations.
%
%
We define $N_d$ stacked 2D heat maps, corresponding to the depth resolution.
The prediction in $(x,y)$ coordinates is performed by applying the Soft-argmax
operation on the averaged heat maps, and the $z$ component is regressed by
applying a one-dimensional Soft-argmax on the volumetric representation
averaged in both $x$ and $y$ dimensions, as depicted in \autoref{fig:3dpose}.
The advantage of splitting the pose prediction into two parts, $(x,y)$ and $z$,
is that we maintain the 2D heat maps as a byproduct, which is useful for
extracting appearance features, as explained in section~\ref{sec:visual}.

With the proposed unified approach, we can train the network with mixed 2D and
3D data. For the first case, only the gradients corresponding to $(x,y)$ are
backpropagated.
As a result, the network can be jointly trained with high precise 3D data from
motion capture systems and very challenging still images collected in
outdoor environments, which are usually manually annotated.

\section{Human action recognition}
\label{sec:action-recognition}

One of the most important advantages in our proposed method is the ability to
integrate high level pose information with low level visual features in
a multitask framework.
This characteristic allows to share the network entry flow for both pose
estimation and visual features extraction.
Additionally, the visual features are trained using both action sequences and
still images captured ``in the wild'', which have been proven as a very efficient
way to learn robust visual representations.

As shown on \autoref{fig:intro}, the proposed action recognition approach is
divided into two parts, one based on a sequence of body joints coordinates,
which we call \textit{pose-based recognition}, and the other based on a
sequence of visual features, which we call \textit{appearance-based
recognition}.
The result of each part is combined to estimate the final action label.
In this section, we give a detailed explanation about each recognition branch,
as well as how we extend single frame pose estimation to extract temporal
information from a sequence of frames.

\subsection{Pose-based recognition}

In order to explore the high level information encoded with body joint
positions, we convert a sequence of $\T$ poses with $\NJoints{}$ joints each
into an image-like representation.
We choose to encode the temporal dimension as the vertical axis, the joints as
the horizontal axis, and the coordinates of each point ($(x,y)$ for 2D,
$(x,y,z)$ for 3D) as the channels.
With this approach, we can use classical 2D convolutions to extract
patterns directly from a temporal sequence of body joints.
Since the pose estimation method is based on still images, we use a time
distributed abstraction to process a video clip, which is a straightforward
technique to handle both single images and video sequences.

\begin{figure}[htbp]
  \centering
  \includegraphics[width=8.00cm]{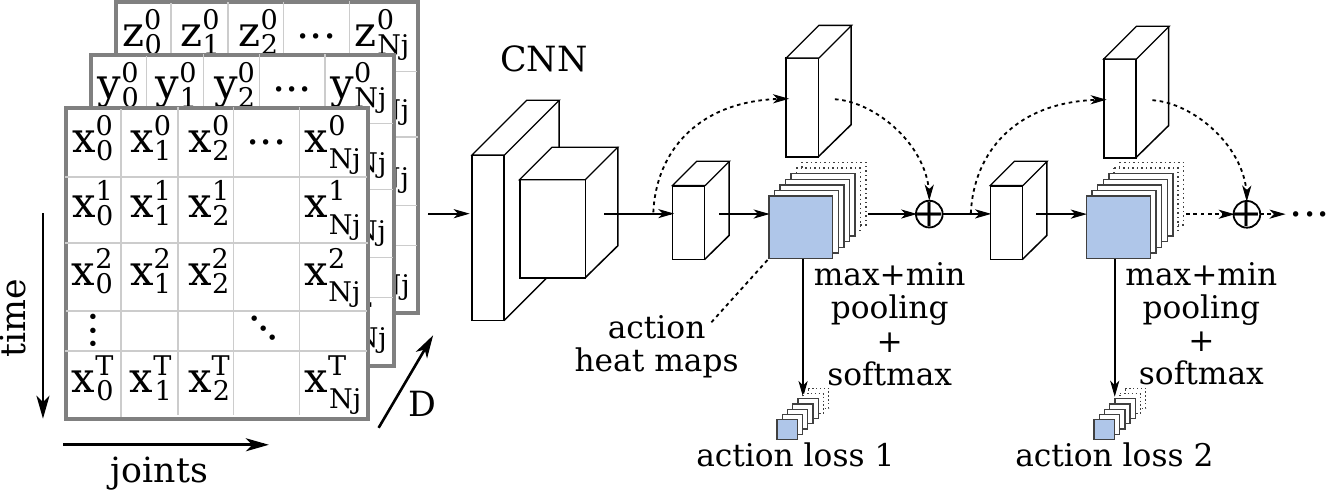}
  \caption{
    Representation of the architecture for action recognition from a sequence of $\T{}$ frames
    of $\NJoints{}$ body joints. The $z$ coordinates are used for 3D action recognition
    only.
    The same architecture is used for appearance-based
    recognition, except that the input are the appearance features instead
    of body joints.
  }
  \label{fig:arposenet}
\end{figure}

We propose a fully convolutional neural network to extract features from input
poses and to produce \textit{action heat maps} as shown on
\autoref{fig:arposenet}.
The idea is that for actions depending only on few body joints, such as
\textit{shaking hands}, fully-connected layers will require zeroing non-related
joints, which is a very difficult learning problem. On the contrary, 2D
convolutions enforce this sparse structure without manually choosing joints and
are thus easier to learn.
Furthermore, different joints have very different coordinates variations and
a filter matching, \eg, hand patterns will not respond to feet patterns equally.
Such patterns are then combined in subsequent layers in order to produce more
discriminative activations until we obtain action maps with a depth equals to
the number of actions.

To produce the output probability of each action for a video clip,
a pooling operation on the action maps has to be performed.
In order to be more sensitive to the strongest responses for each action, we
use the \textit{max plus min} pooling followed by a Softmax activation.
Additionally, inspired by the human pose regression method, we refine
predictions by using a stacked architecture with intermediate supervision
in $\K{}$ prediction blocks. The action heat maps from each prediction block
are then re-injected into the next action recognition block.

\subsection{Appearance-based recognition}
\label{sec:visual}

\begin{figure}[htbp]
  \centering
  \includegraphics[width=7.20cm]{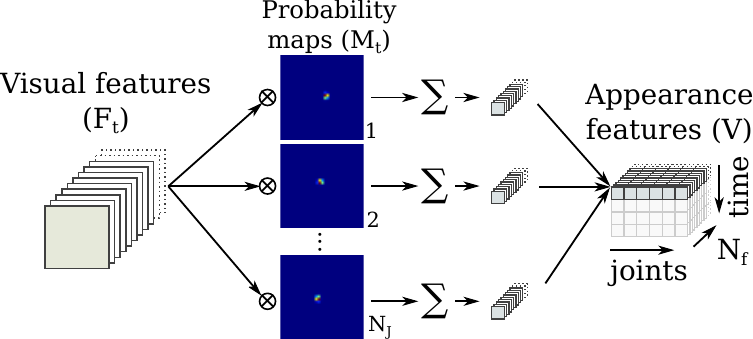}
  \caption{
    Appearance features extraction from low level visual features
    and body parts probability maps for a single frame. For a sequence of $\T{}$
    frames, the appearance features are stacked vertically producing a
    tensor where each line corresponds to one input frame.
  }
  \label{fig:appearancefeat}
\end{figure}

%
The appearance based part is similar to the pose based part, with the difference that it relies on local appearance features instead of joint coordinates.
In order to extract localized appearance features, we multiply the tensor of
visual features $\F{}_t\in\mathbb{R}^{{W_f}\times{H_f}\times{\NFeat{}}}$ obtained at the end of the global entry flow by the
probability maps $\M{}_t\in\mathbb{R}^{{W_f}\times{H_f}\times{\NJoints{}}}$ obtained at the end of the pose estimation part,
where ${W_f}\times{H_f}$ is the size of the feature maps, $\NFeat{}$ is the
number of features, and $\NJoints{}$ is the number of joints.
Instead of multiplying each value individually as in the Kronecker
product, we multiply each channel, resulting in a tensor of size
$\mathbb{R}^{{W_f}\times{H_f}\times{\NJoints{}}\times{\NFeat{}}}$.
Then, the spacial dimensions are collapsed by a sum, resulting in the
appearance features for time $t$ of size $\mathbb{R}^{\NJoints{}\times\NFeat{}}$.
For a sequence of frames, we concatenate each appearance features for
$t=\{0,1,\dots,\T{}\}$ resulting in the video clip appearance features
$\V{}\in\mathbb{R}^{\T{}\times\NJoints{}\times\NFeat{}}$.
To clarify the above appearance features extraction process, a graphical
representation is shown on \autoref{fig:appearancefeat}.

The appearance features are fed into an action recognition network similar to the pose-based action recognition block presented on \autoref{fig:arposenet} with visual features replacing the coordinates of the body joints. 

We argue that our multitask framework has two benefits for the appearance based part: First, it is very computationally efficient since most part of the computations are shared.
Second, the extracted visual features are more robust since they are trained simultaneously for different tasks and on different datasets.

\subsection{Action aggregation}

Some actions are hard to be distinguished from others only by the high level
pose representation. For example, the actions \textit{drink water} and
\textit{make a phone call} are very similar if we take into account only the
body joints, but are easily separated if we have the visual information
corresponding to the objects cup and phone.
On the other hand, other actions are not directly related to visual information
but with body movements, like \textit{salute} and \textit{touch chest}, and in
that case the pose information can provide complementary information.

In order to explore the contribution from both pose and appearance models,
we combine the respective predictions using a fully-connected layer
with Softmax activation, which gives the final prediction of our model.


\begin{figure*}[t!]
  \centering
    \includegraphics[width=0.11\textwidth]{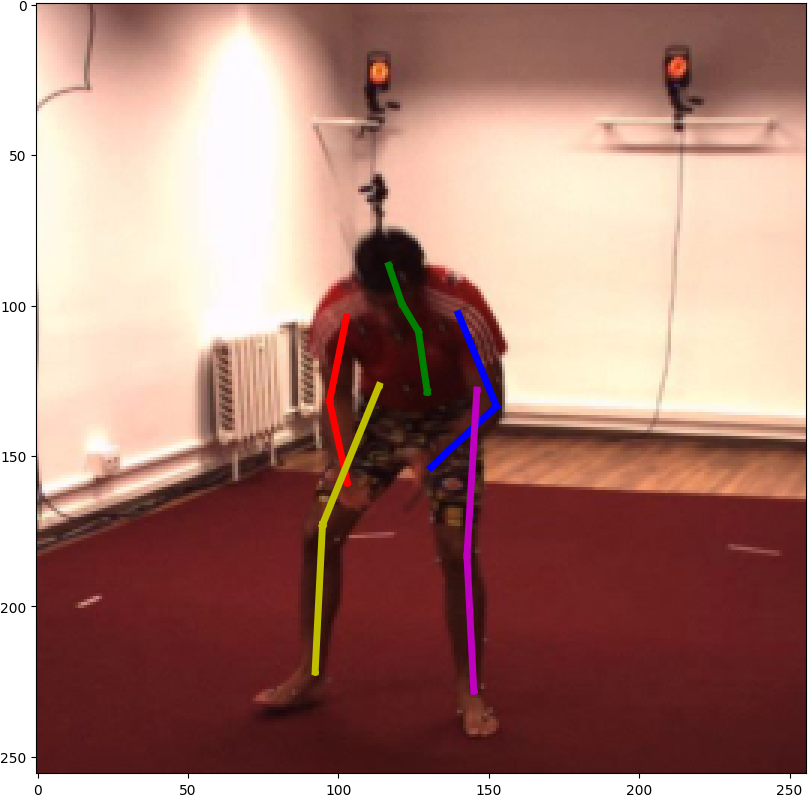}
    \includegraphics[width=0.13\textwidth]{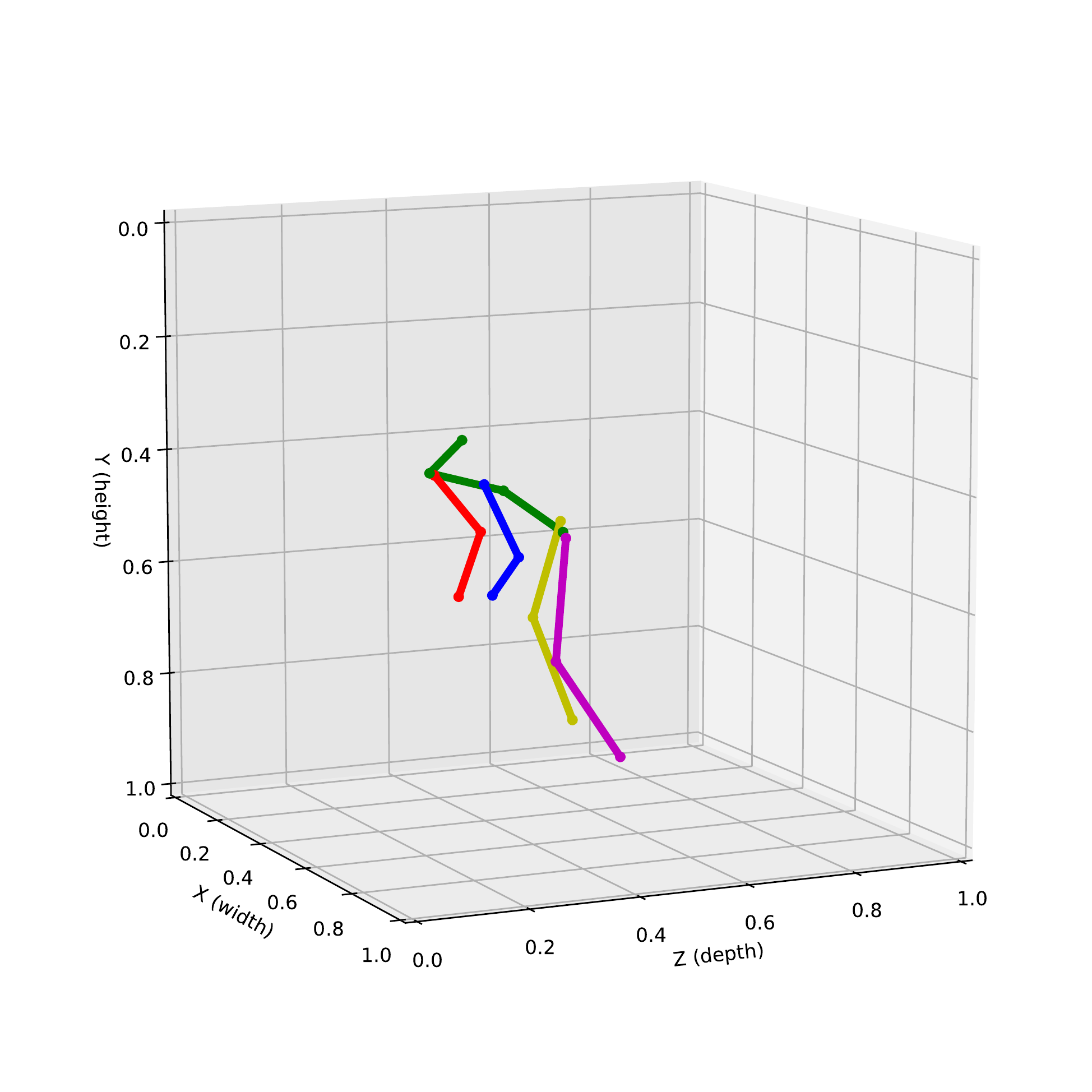}
    \includegraphics[width=0.11\textwidth]{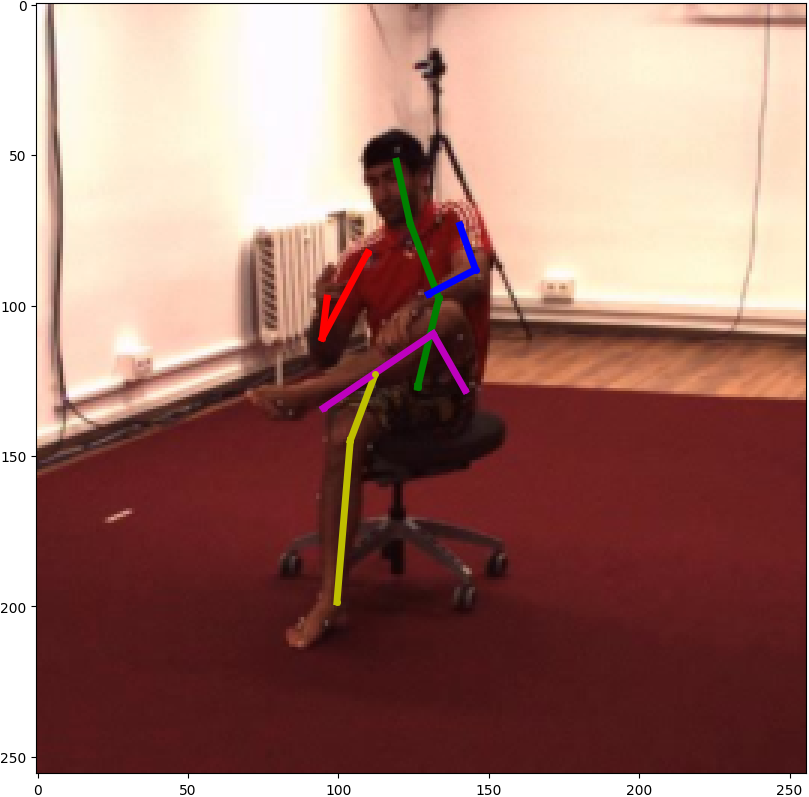}
    \includegraphics[width=0.13\textwidth]{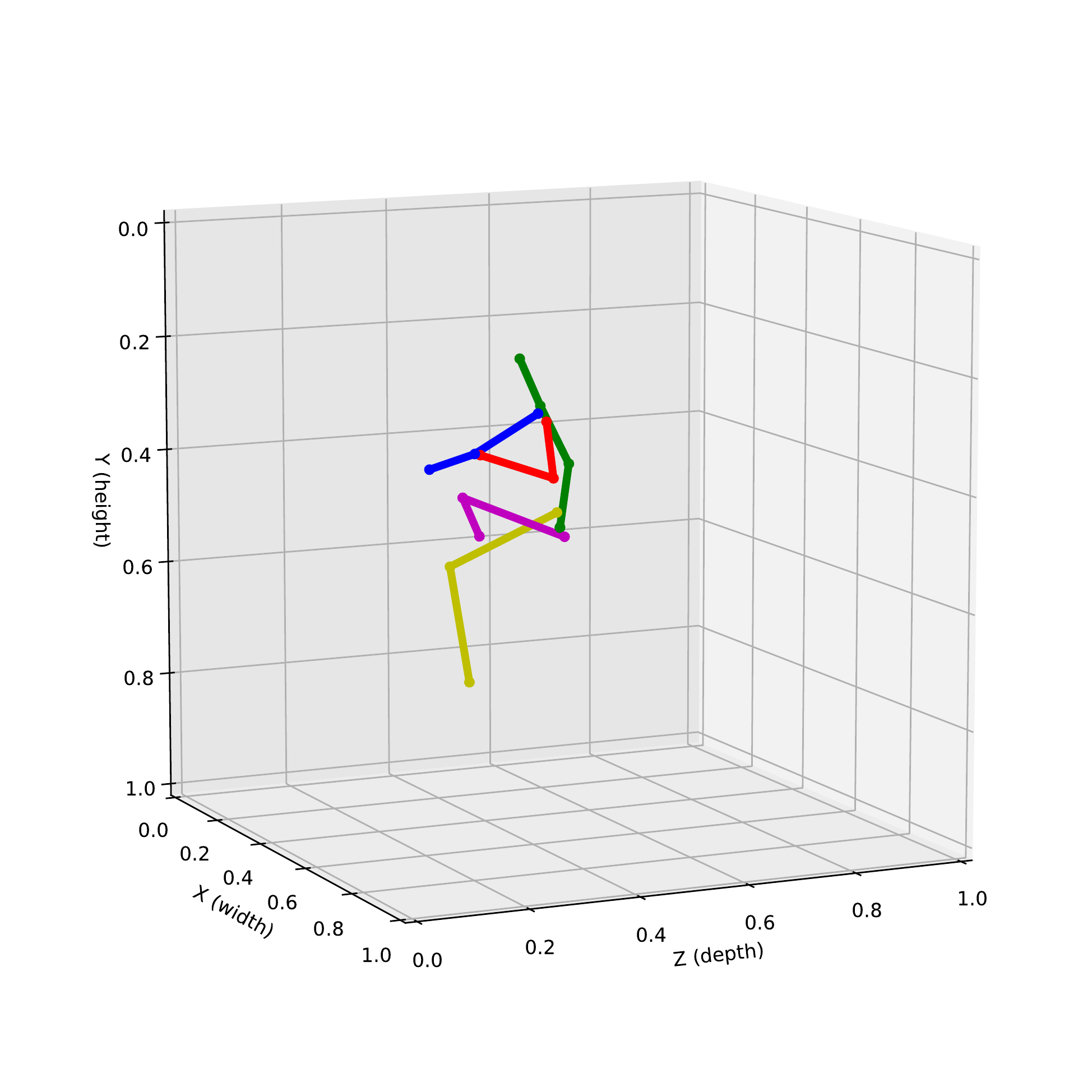}
    \includegraphics[width=0.11\textwidth]{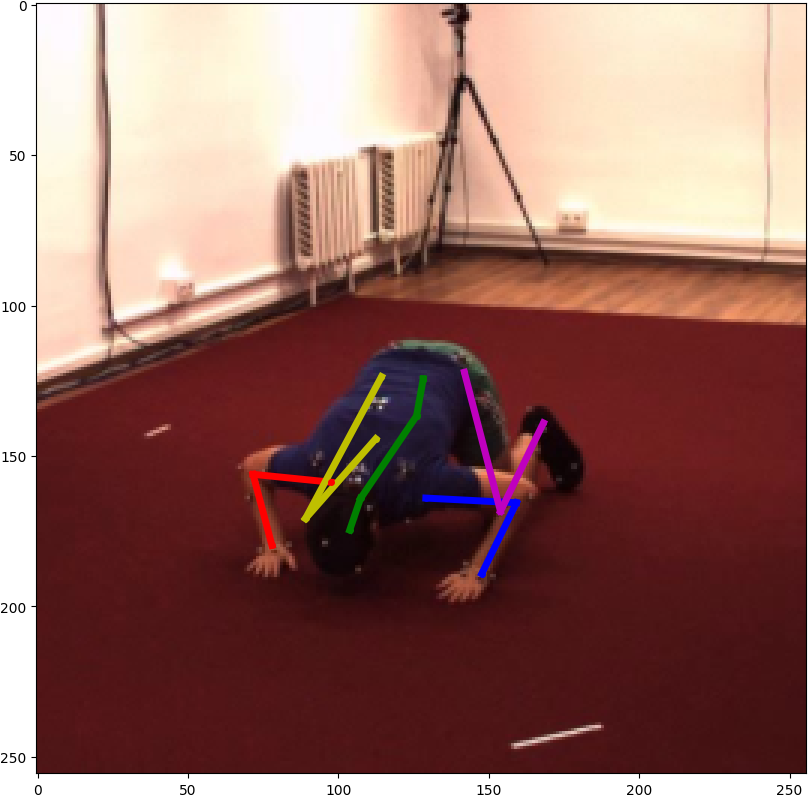}
    \includegraphics[width=0.13\textwidth]{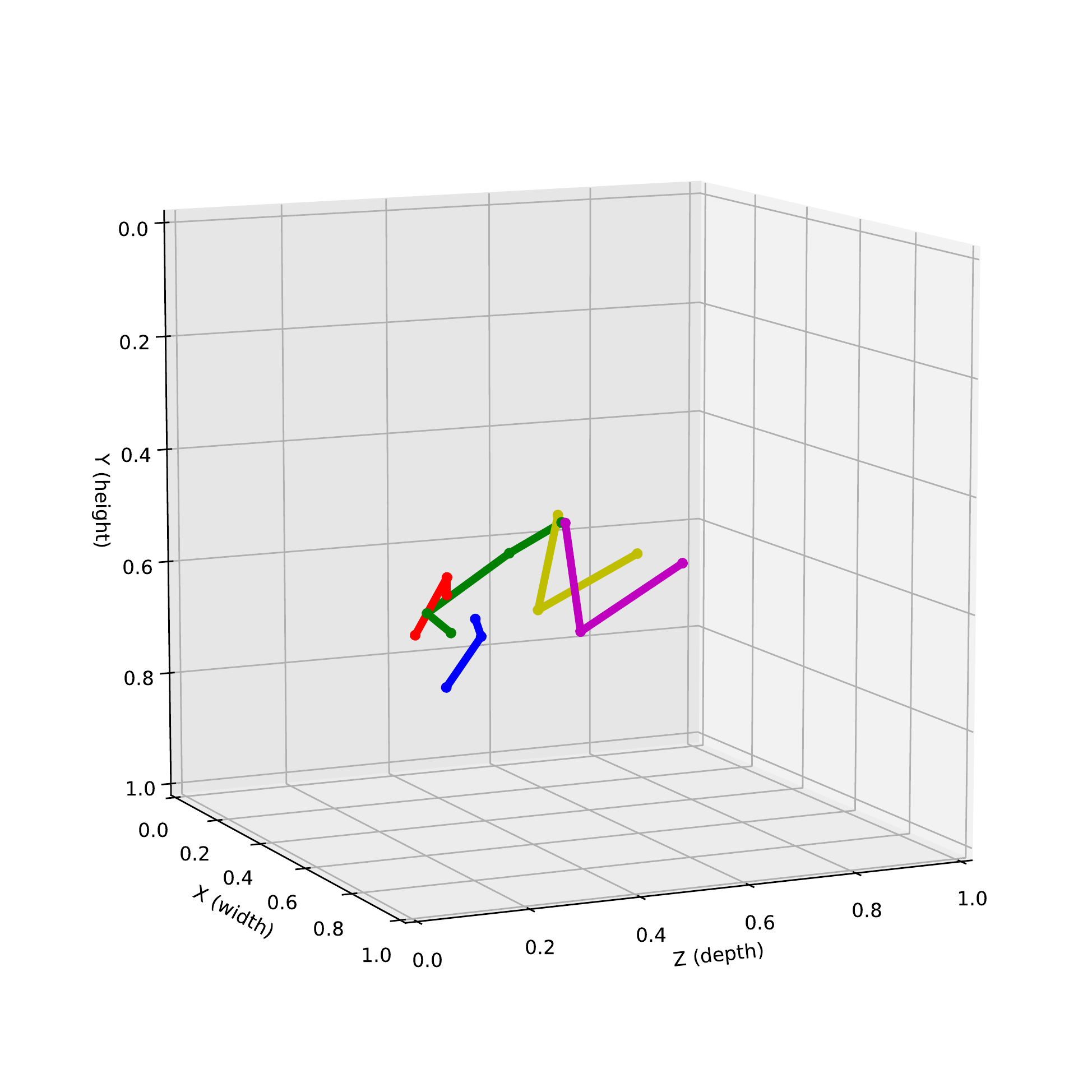}
    \includegraphics[width=0.11\textwidth]{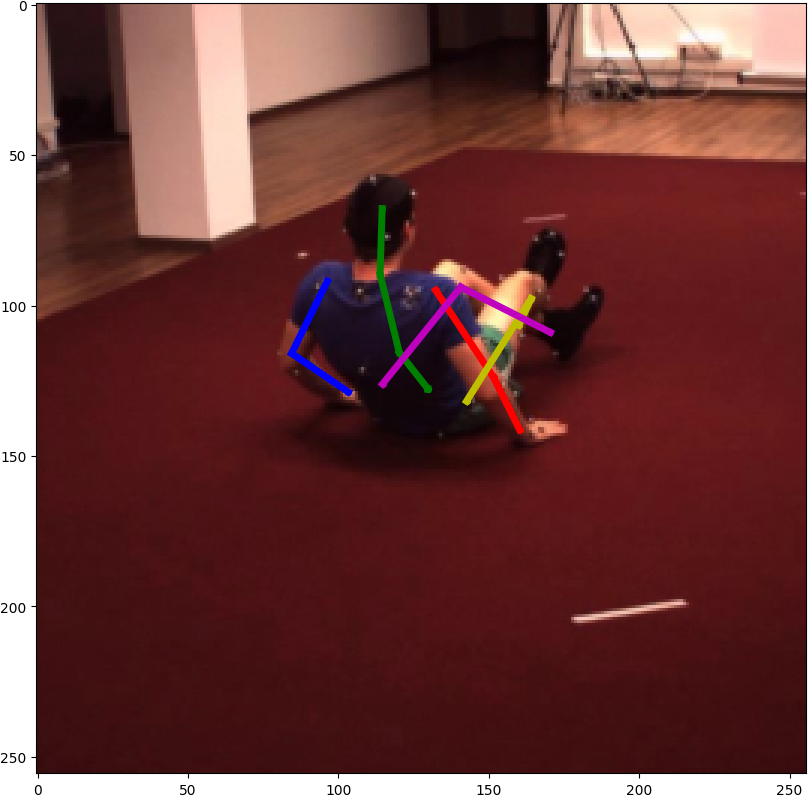}
    \includegraphics[width=0.13\textwidth]{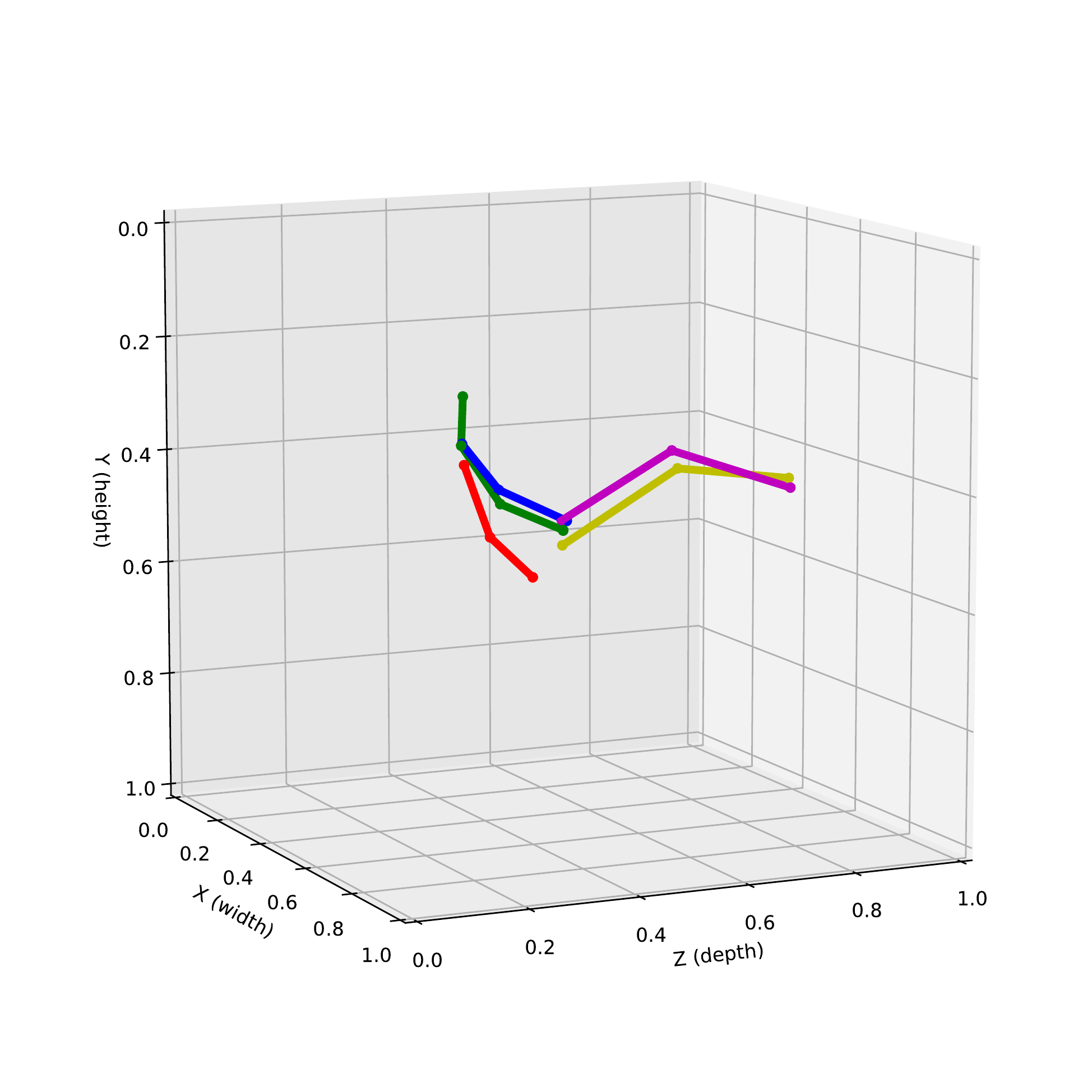}\vspace{-0.25cm}
    \includegraphics[width=0.11\textwidth]{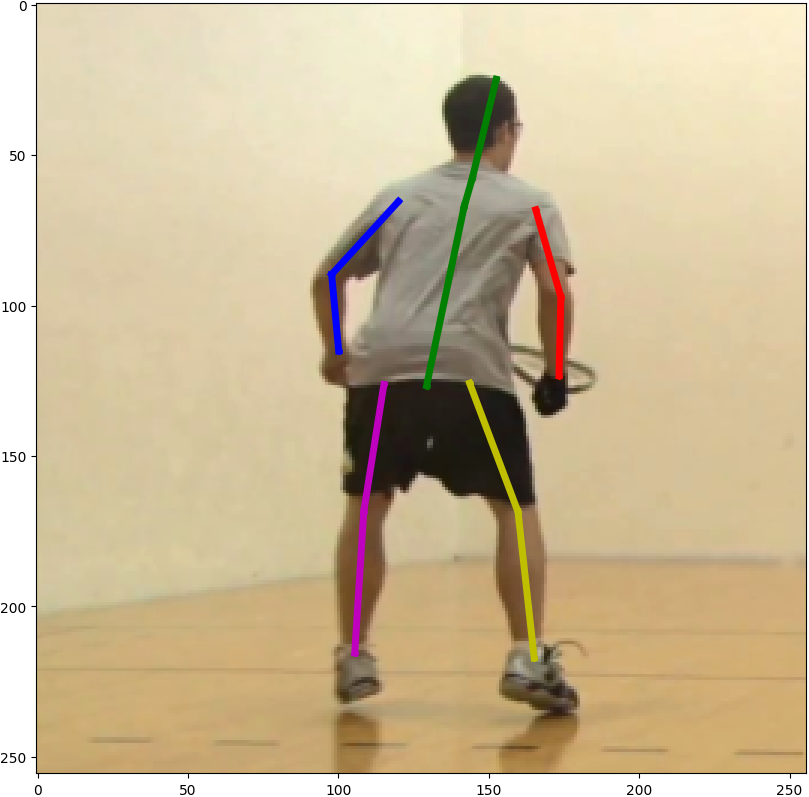}
    \includegraphics[width=0.13\textwidth]{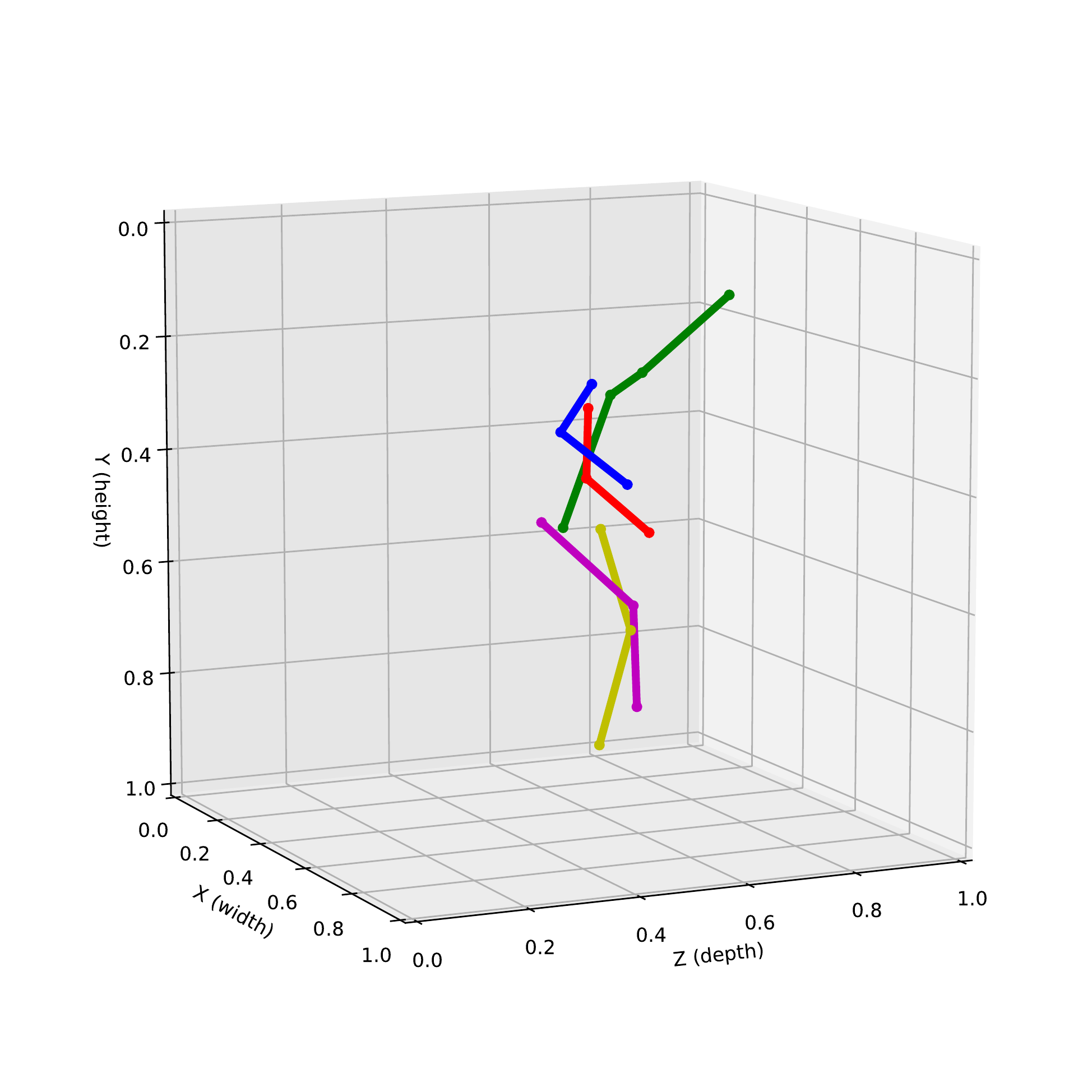}
    \includegraphics[width=0.11\textwidth]{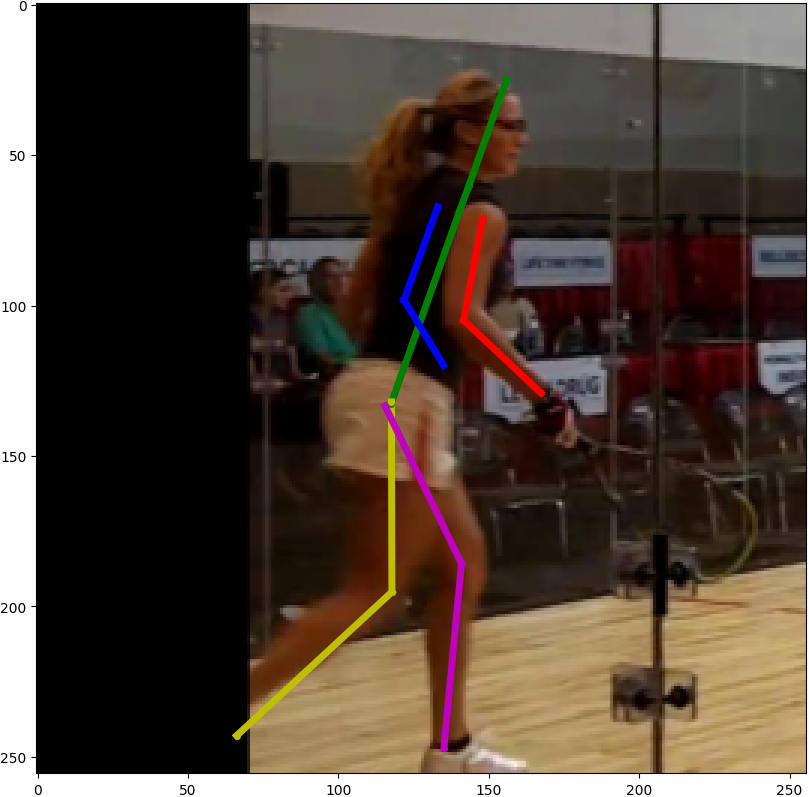}
    \includegraphics[width=0.13\textwidth]{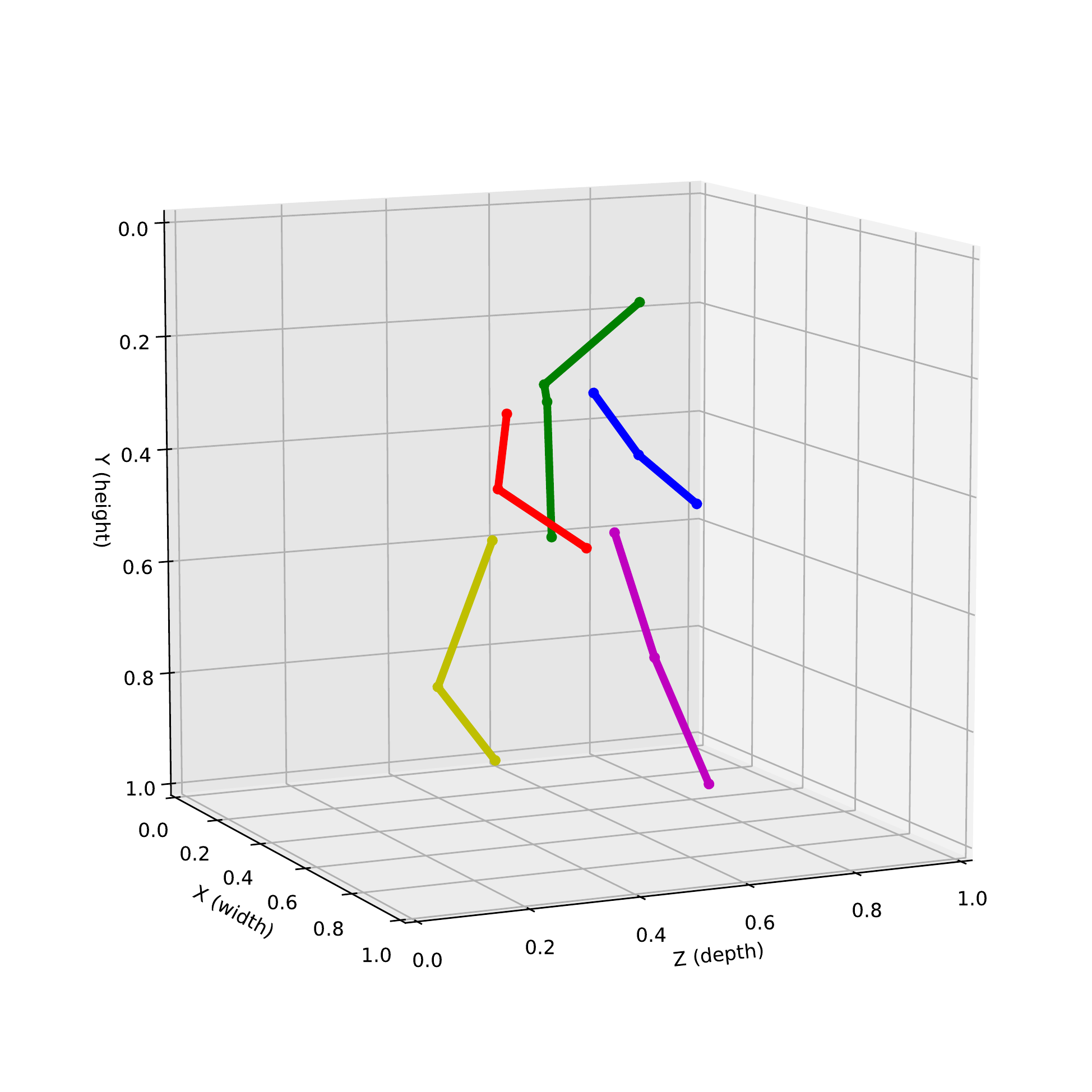}
    \includegraphics[width=0.11\textwidth]{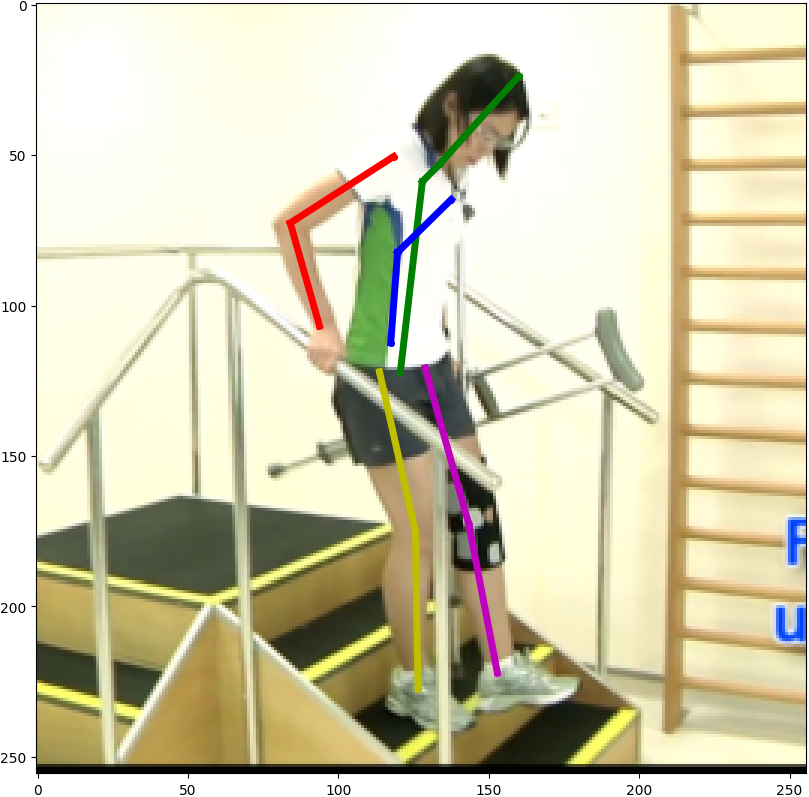}
    \includegraphics[width=0.13\textwidth]{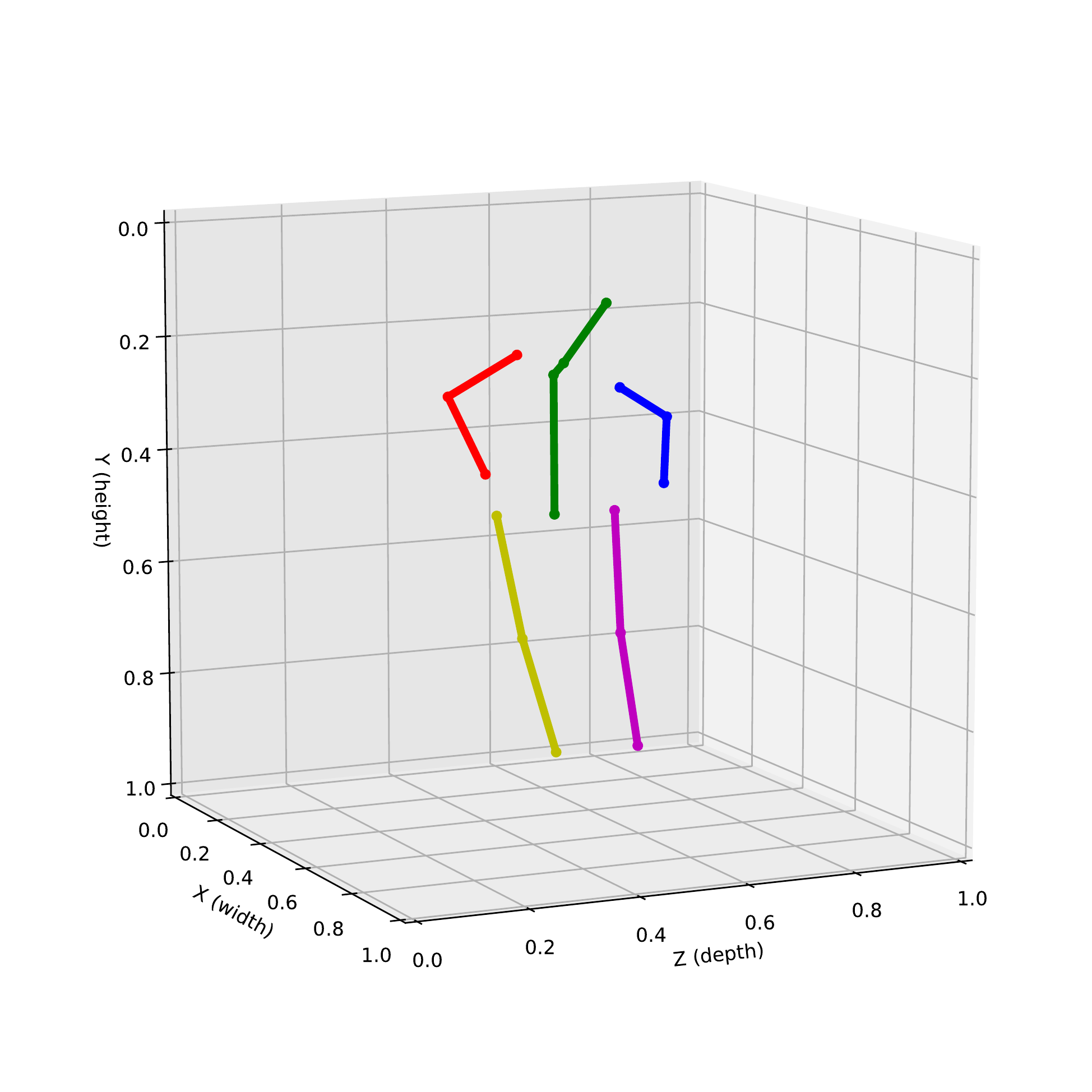}
    \includegraphics[width=0.11\textwidth]{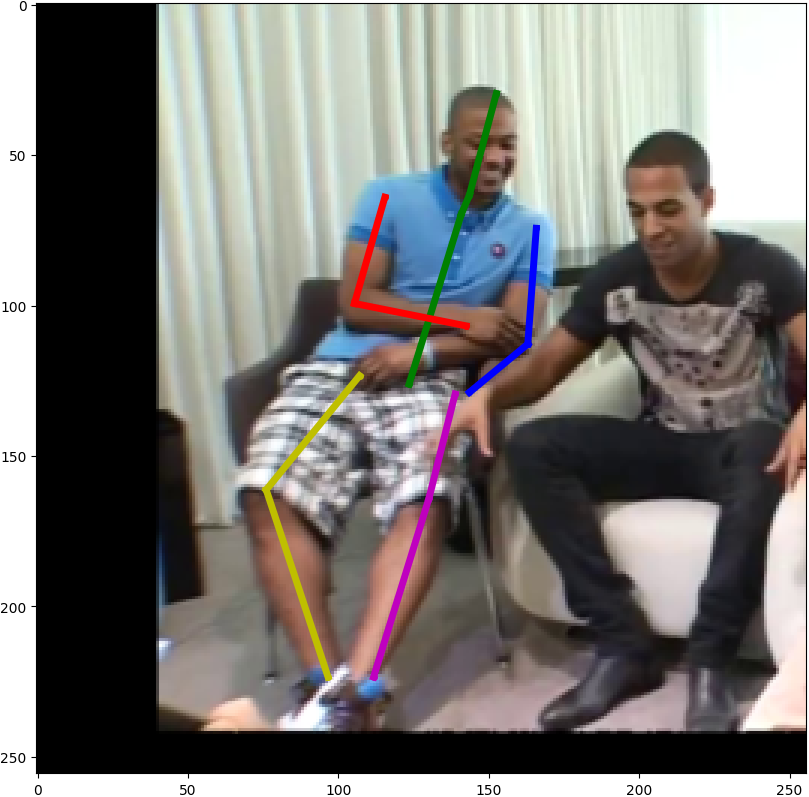}
    \includegraphics[width=0.13\textwidth]{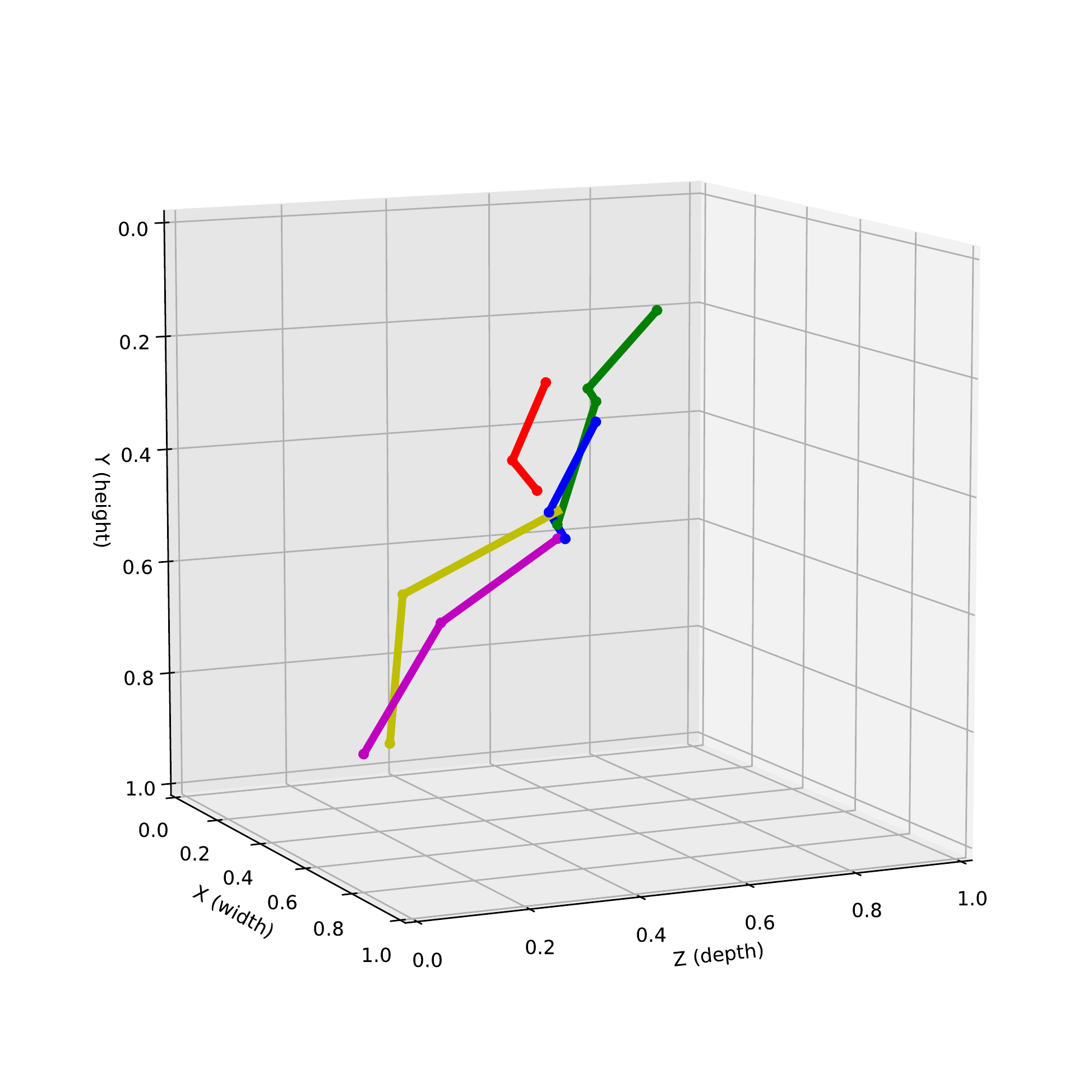}
  \caption{
    Predicted 3D poses from Human3.6M (top row) and MPII (bottom row) datasets.
  }
  \label{fig:3dposes}
\end{figure*}

\section{Experiments}
\label{sec:experiments}

In this section we present the experimental evaluation of our method in four
different categories using four challenging datasets.
We show the robustness and the flexibility of our proposed multitask approach.
The four categories are divided into two problems: human pose estimation and
action recognition. For both cases, we evaluate our approach on 2D and 3D
scenarios.

\subsection{Datasets}

We evaluate our method on four different datasets: on MPII~\cite{andriluka14cvpr}
and on Human3.6M~\cite{h36m_pami} for respectively 2D and 3D pose estimation, 
and on Penn Action~\cite{Zhang_ICCV_2013} and NTU RGB+D~\cite{Shahroudy_2016_CVPR}
for 2D and 3D action recognition, respectively.
The characteristics of each dataset are given as follows.

\vspace{0.05in}\noindent
\textbf{MPII Human Pose Dataset}.
The MPII dataset for single person pose estimation is composed of about 25K
images of which 15K are training samples, 3K are validation samples and
7K are testing samples (which labels are withheld by the authors).
The images are taken from YouTube videos covering 410 different human
activities and the poses are manually annotated with up to 16 body joints.

\vspace{0.05in}\noindent
\textbf{Human3.6M}.
The Human3.6M~\cite{h36m_pami} dataset is composed by videos with 11 subjects
performing 17 different
activities and 4 cameras with different points of view, resulting in more than
3M frames. For each person, the dataset
provides 32 body joints, from which only 17 are used to compute scores.

\vspace{0.05in}\noindent
\textbf{Penn Action }.
The Penn Action dataset~\cite{Zhang_ICCV_2013} is composed by 2,326 videos in the
wild with 15 different actions, among those ``baseball pitch'', ``bench press'',
``strum guitar'', etc.
The challenge on this dataset is that several body parts are
missing in many actions and the image scales are very disparate from one sample to another.

\vspace{0.05in}\noindent
\textbf{NTU RGB+D}.
The NTU dataset is so far the biggest and a very challenging datasets for
3D action recognition. It is composed of more than 56K videos in Full HD of
60 actions performed by 40 different actors and recorded by 3 cameras in 17
different positioning setups, which results in more than 4M video frames.

\begin{table*}[ht]
  \centering
  \caption{Comparison with previous work on Human3.6M evaluated on the averaged
  joint error (in millimeters) on reconstructed poses.}
  \label{tab:result-h36m}
  \footnotesize
  \begin{tabular}{l|cccccccc}
    \hline
    Methods & Direction & Discuss & Eat & Greet & Phone & Posing & Purchase & Sitting \\ \hline
        \hline
    Pavlakos \etal \cite{Pavlakos_2017_CVPR}    & 67.4 & 71.9 & 66.7 & 69.1 & 71.9 & 65.0 & 68.3 & 83.7 \\
    Mehta \etal \cite{MehtaRCSXT16}\textsuperscript{$\star$} & 52.5 & 63.8 & 55.4 & 62.3 & 71.8 & 52.6 & 72.2 & 86.2 \\
    Martinez \etal \cite{Martinez_2017}         & 51.8 & 56.2 & 58.1 & 59.0 & 69.5 & 55.2 & 58.1 & 74.0 \\
    Sun \etal \cite{Sun_2017_ICCV}              & 52.8 & 54.8 & 54.2 & 54.3 & 61.8 & 53.1 & 53.6 & 71.7 \\ \hline
    \textbf{Ours (single-crop)}                 & \textbf{51.5} & \textbf{53.4} & \textbf{49.0} & \textbf{52.5} & \textbf{53.9} & \textbf{50.3} & 54.4 & \textbf{63.6} \\
    \textbf{Ours (multi-crop + h.flip)}         & \textbf{49.2} & \textbf{51.6} & \textbf{47.6} & \textbf{50.5} & \textbf{51.8} & \textbf{48.5} & \textbf{51.7} & \textbf{61.5} \\
    \hline
    Methods & Sit Down & Smoke & Photo & Wait & Walk & Walk Dog & Walk Pair & \multicolumn{1}{|c}{Average} \\ \hline
        \hline
        
    Pavlakos \etal \cite{Pavlakos_2017_CVPR}    & 96.5 & 71.4 & 76.9 & 65.8 & 59.1 & 74.9 & 63.2 & \multicolumn{1}{|c}{71.9} \\
    Mehta \etal \cite{MehtaRCSXT16}\textsuperscript{$\star$} & 120.0& 66.0 & 79.8 & 63.9 & 48.9 & 76.8 & 53.7 & \multicolumn{1}{|c}{68.6} \\
    Martinez \etal \cite{Martinez_2017}         & 94.6 & 62.3 & 78.4 & 59.1 & 49.5 & 65.1 & 52.4 & \multicolumn{1}{|c}{62.9} \\
    Sun \etal \cite{Sun_2017_ICCV}              & 86.7 & 61.5 & 67.2 & 53.4 & 47.1 & 61.6 & 53.4 & \multicolumn{1}{|c}{59.1} \\ \hline
    \textbf{Ours (single-crop)}                 & \textbf{73.5} & \textbf{55.3} & \textbf{61.9} & \textbf{50.1} & \textbf{46.0} & \textbf{60.2} & \textbf{51.0} & \multicolumn{1}{|c}{\textbf{55.1}} \\
    \textbf{Ours (multi-crop + h.flip)}         & \textbf{70.9} & \textbf{53.7} & \textbf{60.3} & \textbf{48.9} & \textbf{44.4} & \textbf{57.9} & \textbf{48.9} & \multicolumn{1}{|c}{\textbf{53.2}} \\\hline
  \end{tabular}\\
  \small\textsuperscript{$\star$} Method not using ground-truth bounding boxes.
\end{table*}

\subsection{Implementation details}
For the pose estimation task, we train the network using the elastic net loss
function on predicted poses as defined in the equation bellow:
\begin{equation}
  \Loss{}_\Pose{} = \frac{1}{\NJoints{}}%
    \sum_{n=1}^{\NJoints{}}\big(%
    \|\hat{\Pose{}}_n-\Pose{}_n\|_1 +%
    \|\hat{\Pose{}}_n-\Pose{}_n \|_2^2\big),
    \label{eq:l1l2-loss}
\end{equation}
where $\hat{\Pose{}}_n$ and $\Pose{}_n$ are respectively the estimated and the
ground truth positions of the $n^{th}$ joint.
For training, we crop bounding boxes centered on the target person by using
the ground truth annotations or the persons location, when applicable.
For the pose estimation task, on both MPII single person and Human3.6M
datasets it is allowed to use the given persons location on evaluation.
If a given body joint falls outside the cropped bounding box on training,
we set the ground truth visibility flag to zero, otherwise we set it to one.
The ground truth visibility information is used to supervise the predicted
joint visibility vector $\JointProb{}$ with the binary cross entropy loss.
When evaluating the pose estimation task we show the results for
\textit{single-crop} and \textit{multi-crop}. In the first case, one
centered image is used for prediction, and on the second case, multiple
images are cropped with small displacements and horizontal flips
and the final pose is the average prediction.

For the action recognition task, we train the network using the categorical
cross entropy loss.
On training, we randomly select fixed-size clips with $\T{}$ frames from a
video sample.
On test, we report results on \textit{single-clip}
or \textit{multi-clip}. In the first case, we crop a single clip in the
middle of the video. For the second case, we crop multiple clips
temporally spaced of $\T{}/2$ frames from each other. The final scores on multi-clip
is computed by the average result on all clips from one video.
To estimate the bounding box on test, we do an initial pose prediction
using the full images from the first, middle, and last frames of a clip.
Finally, we select the maximum bounding box that encloses all the initially
predicted poses.
Detailed information about the network layers and implementation are given in the supplemental material.

\subsection{Evaluation on pose estimation}

\textbf{2D pose estimation}.
We perform quantitative evaluations of the 2D pose estimation using the
probability of correct keypoints measure
with respect to the head size (PCKh), as shown in \autoref{tab:mpii}.
We are able to recover the results of \cite{Luvizon_2017_CoRR} which is
consistent with the similarity between this method and the 2D pose estimation part of our method.
From the results we can see that the regression method based on
Soft-argmax achieves results very close to the state of the art, specially
when considered the accumulated precision given by the area under the
curve (AUC), and by far the most accurate approach among fully
differentiable methods.

\begin{table}[b]
  \centering
  \caption[MPII results]{Comparison results on MPII for
    single person 2D pose estimation using the PCKh measure with respect
    to 0.2 and 0.5 of the head size. For older results, please refer
    to the MPII Leader Board at \url{http://human-pose.mpi-inf.mpg.de}.
  }
  \label{tab:mpii}
  \footnotesize
  \begin{tabular}{@{}lccccc@{}}
    \hline
    Methods & Year & \begin{tabular}[c]{@{}l@{}}PCKh\\@0.2\end{tabular} & \begin{tabular}[c]{@{}l@{}}AUC\\@0.2\end{tabular} & \begin{tabular}[c]{@{}l@{}}PCKh\\@0.5\end{tabular}& \begin{tabular}[c]{@{}l@{}}AUC\\@0.5\end{tabular} \\ \hline
    \multicolumn{6}{c}{Detection methods}  \\
    Recurrent VGG \cite{Belagiannis_2016} & 2016  & 61.6 & 28.2 & 88.1 & 58.8 \\
    DeeperCut \cite{Insafutdinov_ECCV_2016} & 2016  & 64.0 & 31.7 & 88.5 & 60.8 \\
    Pose Machines \cite{Wei_CVPR_2016} & 2016  & 64.8 & 33.0 & 88.5 & 61.4 \\
    Heatmap regression \cite{Bulat_ECCV_2016} & 2016  & 61.8 & 28.5 & 89.7 & 59.6 \\
    Stacked Hourglass \cite{Newell_ECCV_2016} & 2016  & 66.5 & 33.4 & 90.9 & 62.9 \\
    Fractal NN \cite{Ning2017} & 2017  & --  & --  & 91.2 & 63.6 \\
    Multi-Context Att. \cite{Chu_CVPR_17} & 2017 & 67.8 & \textbf{34.1} & 91.5 & 63.8 \\
    Self Adversarial \cite{ChouCC17} & 2017 & \textbf{68.0} & 34.0 & 91.8 & 63.9 \\
    \footnotesize{Adversarial PoseNet\cite{ChenSWLY17}} & 2017 & --  & --  & 91.9 & 61.6 \\
    \footnotesize{Pyramid Res. Module\cite{Yang_2017_ICCV}} & 2017 & --  & --  & \textbf{92.0} & \textbf{64.2} \\
    \hline
    \multicolumn{6}{c}{Regression methods} \\
    LCR-Net \cite{Rogez_CVPR_2017} & 2017 & --    & --    & 74.2  & -- \\
    \footnotesize{Iter. Error Feedback \cite{Carreira_CVPR_2016}} & 2016 & 46.8 & 20.6 & 81.3 & 49.1 \\
    Compositional Reg.\cite{Sun_2017_ICCV} & 2017  & -- & -- & 86.4 & -- \\
    \textbf{2D Soft-argmax} & & \textbf{67.7} & \textbf{34.9} & \textbf{91.2} & \textbf{63.9} \\ \hline
  \end{tabular}
\end{table}

\textbf{3D pose estimation}.
On Human3.6M, we evaluate the proposed 3D pose regression method
by measuring the mean per joint position error (MPJPE), which is the most
challenging and the most common metric for this dataset.
We followed the common evaluation protocol~\cite{Sun_2017_ICCV, Pavlakos_2017_CVPR, MehtaRCSXT16, Chen_2017_CVPR}
by taking five subjects for training
(S1, S5, S6, S7, S8) and evaluating on two subjects (S9, S11) on one every 64 frames.
For training, we use the data equally balanced as 50\%/50\% from MPII and Human3.6M.
For the multi-crop predictions we use five cropped regions and their corresponding flipped images.
Our results compared to the previous approaches are presented in \autoref{tab:result-h36m} and show that our approach is able to outperform the state of the art by a fair margin.
Qualitative results from our method are shown in \autoref{fig:3dposes}, for both Human3.6M and MPII datasets,
which also demonstrate the capability of our method to generalize 3D pose predictions
from data with only 2D annotated poses.

\subsection{Evaluation on action recognition}

\textbf{2D action recognition}.
We evaluate our action recognition approach on 2D scenario on the Penn Action dataset.
For training the pose estimation part, we use mixed data from
MPII (75\%) and Penn Action (25\%), using 16 body joints.
The action recognition part was trained using video clips composed of $T=16$ frames.
We reached state of the art classification score among methods using RGB and estimated poses.
We also evaluated our method without considering the influence of estimated poses by using
the manually annotated body joints and are also able to improve over the state of the art.
Results are shown in \autoref{tab:pennaction}.

\begin{table}[h]
  \centering
  \caption[PennAction results]{Comparison results on Penn Action for
    2D action recognition. Results given as the percentage of correctly classified actions.
  }
  \label{tab:pennaction}
  \small
  \begin{tabular}{@{}lccccc@{}}
    \hline
    Methods & \begin{tabular}[c]{@{}c@{}}\footnotesize Annot.\\poses\end{tabular} & RGB & \begin{tabular}[c]{@{}c@{}}\footnotesize Optical\\Flow\end{tabular} & \begin{tabular}[c]{@{}c@{}}\footnotesize Estimated\\poses\end{tabular} & Acc. \\ \hline
    Nie \etal \cite{Nie_2015_CVPR}                  & - & X & - & X & 85.5 \\ \hline
    \multirow{2}{*}{Iqbal \etal \cite{Iqbal_2017}}  & - & - & - & X & 79.0 \\
                                                    & - & X & X & X & 92.9 \\ \hline
    \multirow{2}{*}{Cao \etal \cite{Cao_2017}}      & X & X & - & - & 98.1 \\
                                                    & - & X & - & X & 95.3 \\
    \hline
    \multirow{2}{*}{\textbf{Ours}} & X & X & - & - & \textbf{98.6} \\
                                                    & - & X & - & X\textsuperscript{$\star$} & \textbf{97.4} \\ \hline
  \end{tabular} \\
  \small\textsuperscript{$\star$} Using mixed data from PennAction and MPII.
\end{table}

\begin{table}[ht]
  \centering
  \caption[NTU results]{Comparison results on the NTU for 3D action recognition.
  Results given as the percentage of correctly classified actions
  }
  \label{tab:ntu}
  \small
  \begin{tabular}{@{}lcccc@{}}
    \hline
    Methods & \begin{tabular}[c]{@{}c@{}}\footnotesize Kinect\\poses \end{tabular} & RGB & \begin{tabular}[c]{@{}c@{}}\footnotesize Estimated\\poses \end{tabular} & \begin{tabular}[c]{@{}c@{}}Acc. cross\\subject\end{tabular} \\ \hline
    Shahroudy \etal \cite{Shahroudy_2016_CVPR}          & X & - & - & 62.9 \\
    Liu \etal \cite{Liu2016}                            & X & - & - & 69.2 \\
    Song \etal \cite{Song_2017_AAAI}                    & X & - & - & 73.4 \\
    Liu \etal \cite{Liu_2017_CVPR}                      & X & - & - & 74.4 \\
    Shahroudy \etal \cite{Shahroudy2017DeepMF}          & X & X & - & 74.9 \\ \hline
    \multirow{3}{*}{Baradel \etal \cite{baradel2017a}}  & X & - & - & 77.1 \\
                                                        & \textsuperscript{$\star$} & X & - & 75.6 \\
                                                        & X & X & - & 84.8 \\
    \hline
    \multirow{2}{*}{\textbf{Ours}}  & - & X & - & \textbf{84.6} \\
      & - & X & X & \textbf{85.5} \\ \hline
  \end{tabular} \\
  \small\textsuperscript{$\star$} GT poses were used on test to select visual features.
  \vspace{-0.1cm}
\end{table}

\textbf{3D action recognition}.
Since skeletal data from NTU is frequently noisy, we train the pose
estimation part with only 10\% of data from NTU, 45\% from MPII, and 45\% from Human3.6M,
using 20 body joints and video clips of $T=20$ frames.
Our method improves the state of the art on NTU significantly using
only RGB frames and 3D predicted poses, as reported in \autoref{tab:ntu}.
If we consider only RGB frames as input, our method improves over \cite{baradel2017a} by 9.9\%.
To the best of our knowledge, all the previous methods use provided poses
given by Kinect-v2, which are known to be very noisy in some cases.
Although we do not use LSTM like other methods, the temporal information is
well taken into account using convolution. Our results suggest this approach is
sufficient for small video clips as found in NTU.

\textbf{Ablation study}.
We performed varied experiments on NTU to show the contributions
of each component of our methods. As can be seen on \autoref{tab:ablation-ntu},
our estimated poses increase the accuracy by 2.9\% over Kinect poses.
Moreover, the full optimization also improves by 3.3\%, which justify the importance
of a fully differentiable approach. And finally, by averaging results from multiple
video clips we gain 1.1\% more.
We also compared the proposed approach of sequential learning followed by
fine tuning (\autoref{tab:pennaction}) with joint learning pose and action on
PennAction, what result in 97.3\%, only 0.1\% lower than in the previews case.

\begin{figure}[htbp]
  \centering
  \includegraphics[width=8.70cm]{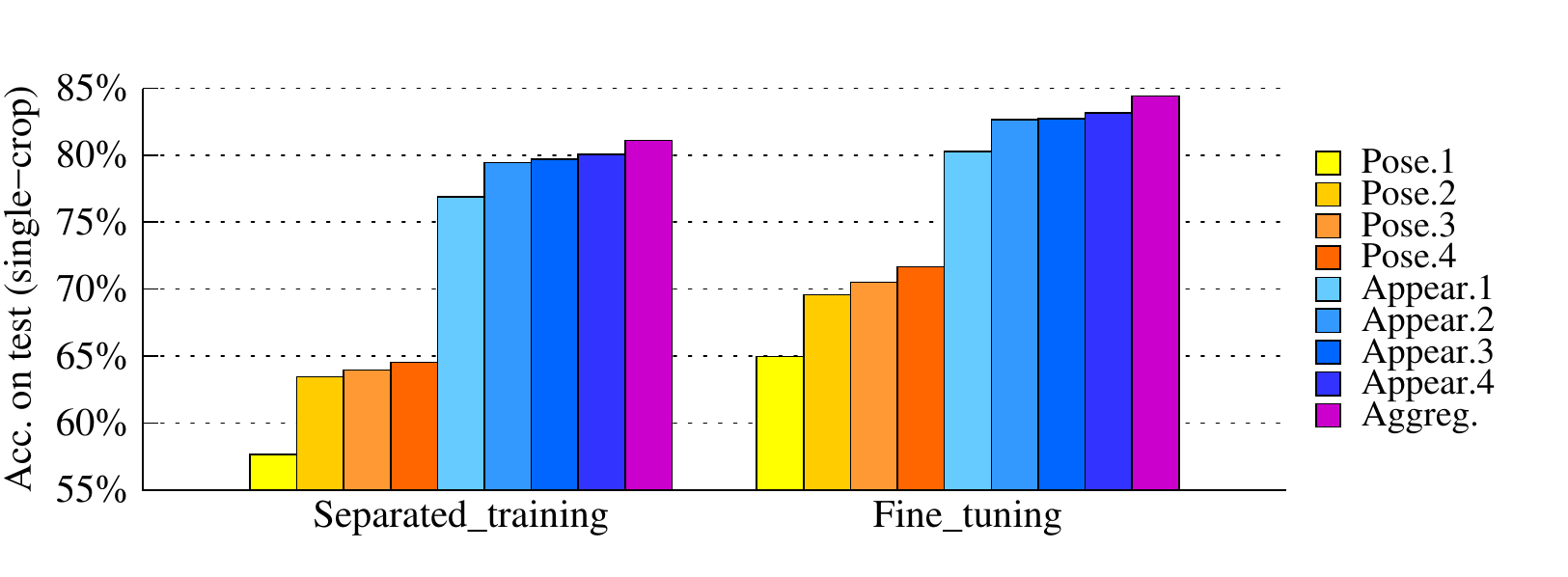}
  \caption{
    Action recognition accuracy on NTU from pose and appearance models in four prediction blocks,
    and with aggregated features, for both separated training and
    full network optimization (fine tuning).
  }
  \label{fig:gain_ar}
\end{figure}

\begin{figure}[htbp]
  \centering
  \includegraphics[width=8.70cm]{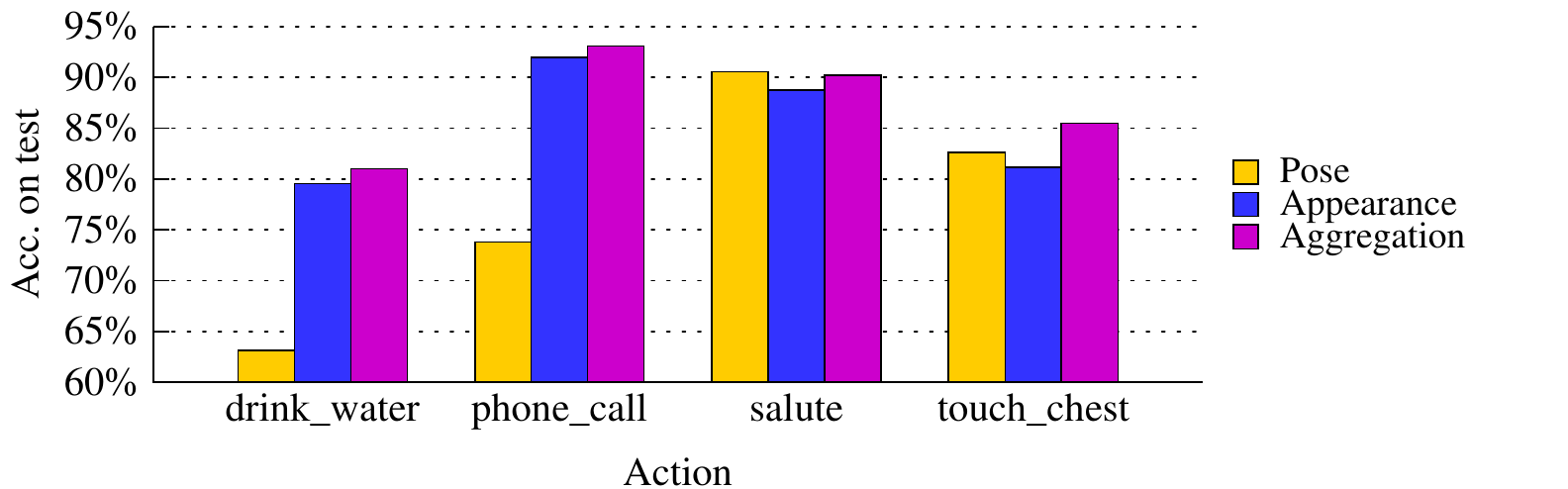}
  \caption{
    Action recognition accuracy on NTU for different action types from pose,
    and appearance models and with aggregated results.
  }
  \label{fig:categorical_ar}
\end{figure}

The effectiveness of our method relies on three main characteristics: 
First, the multiple prediction blocks provide a continuous improvement
on action accuracy, as can be seen on \autoref{fig:gain_ar}.
Second, thanks to our fully differentiable architecture, we can fine tune the
model from RGB frames to predicted actions, which brings a significant gain
in accuracy.
And third, as shown on \autoref{fig:categorical_ar}, the proposed approach
also benefits from complementary appearance and pose information which lead to better classification accuracy once aggregated.

\begin{table}[t]
  \centering
  \caption{Results of our method on NTU considering different approaches. FT: Fine tuning, MC: Multi-clip.}
  \label{tab:ablation-ntu}
  \small
  \begin{tabular}{l|ccc}
    \hline
    Experiments & Pose & \multicolumn{1}{c}{\begin{tabular}[c]{@{}c@{}}Appearance\\ (RGB)\end{tabular}} &
    \multicolumn{1}{c}{Aggregation} \\ \hline
    Kinect poses                                               & 63.3 & 76.4 & 78.2 \\ \hline
    Estimated poses                                            & 64.5 & 80.1 & 81.1 \\ \hline
    \begin{tabular}[c]{@{}l@{}}Est. poses + FT\end{tabular}    & 71.7 & 83.2 & 84.4 \\ \hline
    Est. poses + FT + MC                                       & 74.3 & 84.6 & 85.5 \\ \hline
  \end{tabular}
  \vspace{-0.1cm}
\end{table}

\section{Conclusions}
\label{sec:conclusions}

In this paper, we presented a multitask deep architecture to perform 2D and 3D pose estimation jointly with action recognition.
Our model first predicts the 2D and 3D location of body joints from the raw RGB frames.
These locations are then used to predict the action performed in the video in two different ways: using semantic information by leveraging the temporal evolution of body joint coordinates and using visual information by performing an attention based pooling on human body parts.
Heavy sharing of weights and features in our model allows us to solve four different tasks - 2D pose estimation, 3D pose estimation, 2D action recognition, 3D action recognition - with a single model very efficiently compared to dedicated approaches.
We performed extensive experiments that show our approach is able to equal or even outperform dedicated approaches on all these tasks.
%

\section{Acknowledgements}

This work was partially founded by CNPq (Brazil) - Grant 233342/2014-1.

\section*{Appendix A: Network architecture}

\begin{figure}[htbp]
  \centering
  \includegraphics[width=0.22\textwidth]{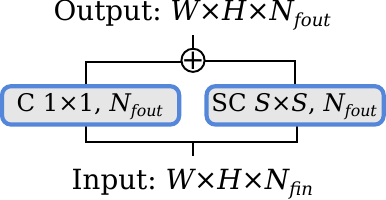}\hspace{0.2cm}
  \includegraphics[width=0.22\textwidth]{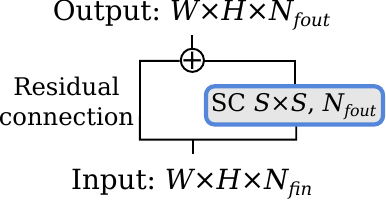}
  \caption{
    Separable residual module (SR) based on depth-wise separable convolutions
    (SC) for
    $N_{fin} \neq N_{fout}$ (left), and $N_{fin} = N_{fout}$ (right), where
    $N_{fin}$ and $N_{fout}$ are the input and output features size,
    $W\times{H}$ is the feature map resolution, and $S\times{S}$ is
    the size of the filters, usually $3\times3$ or $5\times5$.
    C: Simple 2D convolution.
  }
  \label{fig:sep-residual}
\end{figure}

\begin{figure}[htbp]
  \centering
  \includegraphics[width=0.21\textwidth]{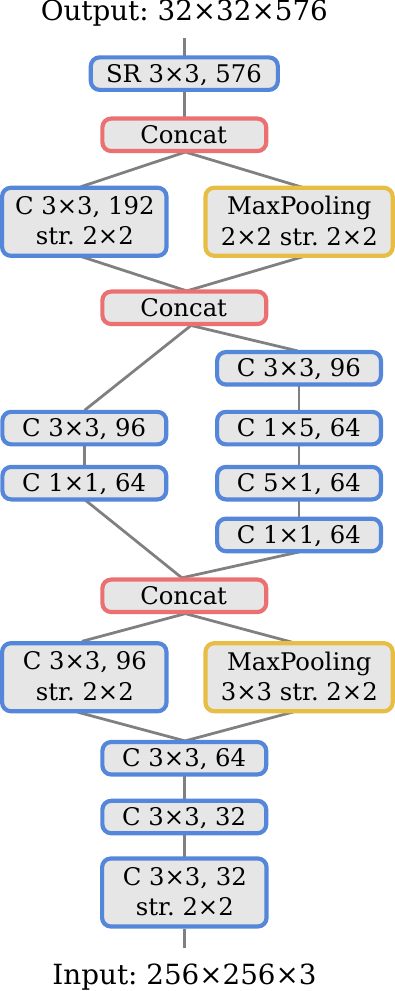}
  \caption{
    Shared network (entry flow) based on Inception-V4.
    C: Convolution, SR: Separable residual module.
  }
  \label{fig:net-stem}
\end{figure}

\begin{figure}[htbp]
  \centering
  \includegraphics[width=0.35\textwidth]{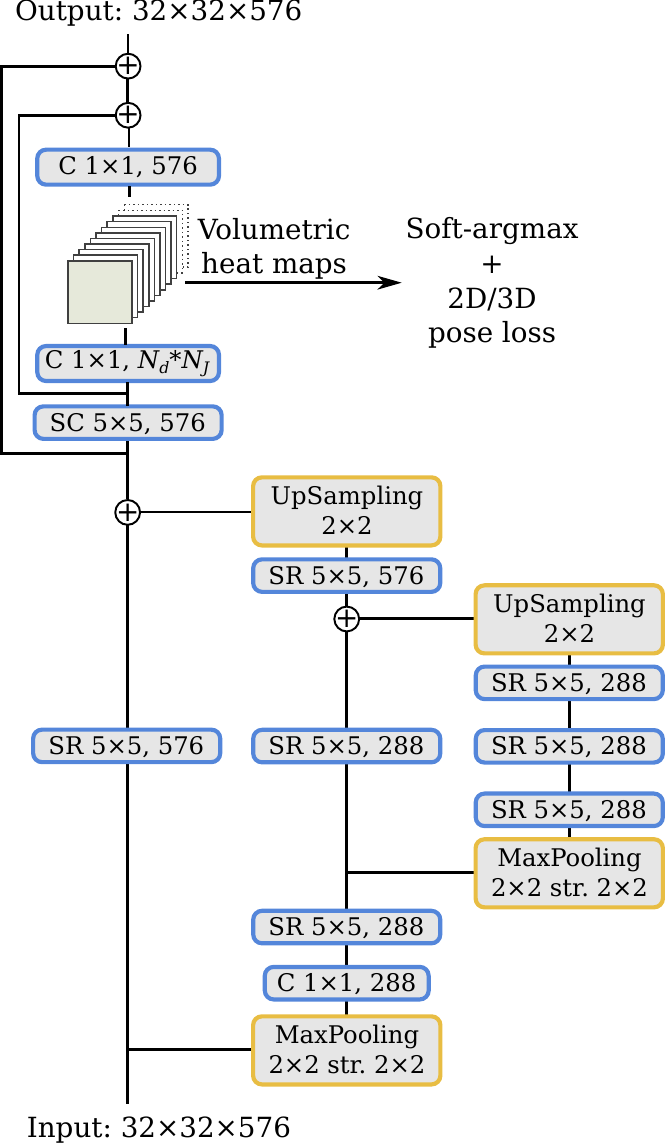}
  \caption{
    Prediction block for pose estimation, where $N_d$ is the number of depth heat maps
    per joint and $\NJoints{}$ is the number of body joints.
    C: Convolution, SR: Separable residual module.
  }
  \label{fig:net-pred-pose}
\end{figure}

\begin{figure}[htbp]
  \centering
  \includegraphics[width=0.35\textwidth]{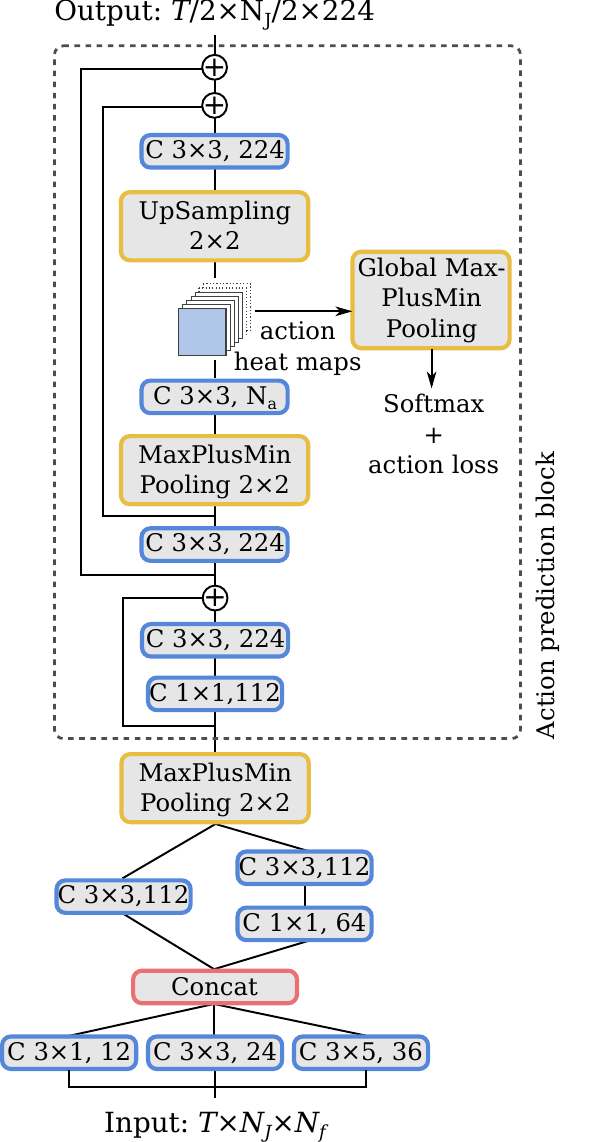}
  \caption{
    Network architecture for action recognition. The action prediction blocks can be repeated $K$ times.
    The same architecture is used for pose and appearance recognition, except that
    for pose, each convolution uses half the number of features showed here.
    $T$ corresponds the number of frames and $N_a$ is the number of actions.
  }
  \label{fig:net-ar}
\end{figure}

In our implementation of the proposed approach, we divided the network
architecture into four parts: the \textit{multitask stem}, the
\textit{pose estimation model}, the \textit{pose recognition model}, and
the \textit{appearance recognition model}.
We use depth-wise separable convolutions as depicted in
\autoref{fig:sep-residual}, batch normalization and ReLu activation.
The architecture of the multitask stem is detailed in \autoref{fig:net-stem}.
Each pose estimation prediction block is implemented as a multi-resolution
CNN, as presented in \autoref{fig:net-pred-pose}. We use $N_d$ = 16 heat maps
for depth predictions.
The CNN architecture for action recognition is detailed in \autoref{fig:net-ar}.

\section*{Appendix B: Training parameters}

\begin{table*}[!htb]
  \centering
  \caption{Our results on averaged joint error on reconstructed poses for 3D pose estimation on Human3.6
  considering single dataset training (Human3.6M only) and mixed data (Human3.6M + MPII).
  SC: Single-crop, MC: Multi-crop.
  }
  \label{tab:add-result-h36m}
  \small
  \begin{tabular}{l|cccccccc}
    \hline
    Methods & Direction & Discuss & Eat & Greet & Phone & Posing & Purchase & Sitting \\ \hline
        \hline
    \textbf{Human3.6 only - SC}     & 64.1 & 66.3 & 59.4 & 61.9 & 64.4 & 59.6 & 66.1 & 78.4 \\
    \textbf{Human3.6 only - MC}     & 61.7 & 63.5 & 56.1 & 60.1 & 60.0 & 57.6 & 64.6 & 75.1 \\
    \textbf{Human3.6 + MPII - SC}   & 51.5 & 53.4 & 49.0 & 52.5 & 53.9 & 50.3 & 54.4 & 63.6 \\
    \textbf{Human3.6 + MPII - MC}   & 49.2 & 51.6 & 47.6 & 50.5 & 51.8 & 48.5 & 51.7 & 61.5 \\
    \hline
    Methods & Sit Down & Smoke & Photo & Wait & Walk & Walk Dog & Walk Pair & \multicolumn{1}{|c}{Average} \\ \hline
        \hline
    
    \textbf{Human3.6 only - SC}     & 102.1& 67.4 & 77.8 & 59.3 & 51.5 & 69.7 & 60.1 & \multicolumn{1}{|c}{67.3} \\
    \textbf{Human3.6 only - MC}     & 95.4 & 63.4 & 73.3 & 57.0 & 48.2 & 66.8 & 55.1 & \multicolumn{1}{|c}{63.8} \\
    \textbf{Human3.6 + MPII - SC}   & 73.5 & 55.3 & 61.9 & 50.1 & 46.0 & 60.2 & 51.0 & \multicolumn{1}{|c}{55.1} \\
    \textbf{Human3.6 + MPII - MC}   & 70.9 & 53.7 & 60.3 & 48.9 & 44.4 & 57.9 & 48.9 & \multicolumn{1}{|c}{53.2} \\\hline
  \end{tabular}
\end{table*}

In order to merge different datasets, we convert the poses to a common
layout, with a fixed number of joints equal to the dataset with more joints.
For example, when merging the datasets Human3.6M and MPII, we use all the 17
joints in the first dataset and include one joint on MPII.
All the included joints have an invalid value that is not taken into
account in the loss function.
Additionally, we use and alternated human pose layout, similar to the layout
from the Penn Action dataset, which experimentally lead to better scores
on action recognition.

We optimize the pose regression part using the RMSprop optimizer with initial
learning rate of 0.001, which is reduced by a factor of 0.2 when validation score
plateaus, and batches of 24 images.
For the action recognition task, we train both pose and appearance models
simultaneously using a pre-trained pose estimation model with weights
initially frozen. In that case, we use a classical SGD optimizer with Nesterov
momentum of 0.98 and initial learning rate of 0.0002, reduced by a factor of 0.2 when
validation plateaus, and batches of 2 video clips.
When validation accuracy stagnates, we divide the final learning rate by 10 and
fine tune the full network for more 5 epochs.
When reporting only pose estimation scores, we use eight prediction blocks
($\K{}=8$), and for action recognition, we use four prediction blocks ($\K{}=4$).
For all experiments, we use cropped RGB images of size $256\times256$.
We augment the training data by performing random rotations from $-45^{\circ}$
to $+45^{\circ}$, scaling from $0.7$ to $1.3$, vertical and horizontal
translations respectively from $-40$ to $+40$ pixels, video subsampling by a factor from 1 to 3,
and random horizontal flipping.

\section*{Appendix C: Additional experiments}

In order to show the contribution of multiple datasets in training, we show in
\autoref{tab:add-result-h36m} additional results on 3D pose estimation using
Human3.6M only and Human3.6M + MPII datasets for training.

{
  \small
  \bibliographystyle{ieee}
  \bibliography{references}
}

\end{document}